\documentclass{article}

     \usepackage[preprint]{neurips_2021}

\usepackage[utf8]{inputenc} % allow utf-8 input
\usepackage{natbib}
\usepackage[T1]{fontenc}    % use 8-bit T1 fonts
\usepackage{hyperref}       % hyperlinks
\usepackage{url}            % simple URL typesetting
\usepackage{booktabs}       % professional-quality tables
\usepackage{amsfonts}       % blackboard math symbols
\usepackage{nicefrac}       % compact symbols for 1/2, etc.
\usepackage{microtype}      % microtypography
\usepackage{xcolor}         % colors
\usepackage{algorithm} 
\usepackage{algpseudocode}
\usepackage{commath}
\usepackage{amsthm, amssymb}
\usepackage{graphicx}
\usepackage{verbatim}
\usepackage{bbm}
\DeclareMathOperator*{\argmax}{arg\,max}

\newtheorem{theorem}{Theorem}[section]

\newtheorem{lemma}[theorem]{Lemma}
\newtheorem{prop}{Proposition}[section]
\newtheorem{assumption}{Assumption}[section]

\title{Computationally Efficient High-Dimensional Bayesian Optimization via Variable Selection}

\author{%
  Yihang Shen \\
  Department of Computational Biology\\
  Carnegie Mellon University\\
  Pittsburgh, PA 15213 \\
  \texttt{yihangs@andrew.cmu.edu} \\
   \And
   Carl Kingsford\thanks{Corresponding author} \\
   Department of Computational Biology \\
   Carnegie Mellon University\\
   Pittsburgh, PA 15213 \\
   \texttt{carlk@cs.cmu.edu} \\
}

\begin{document}

\maketitle

\begin{abstract}
  Bayesian Optimization (BO) is a method for globally optimizing black-box functions. While BO has been successfully applied to many scenarios, developing effective BO algorithms that scale to functions with high-dimensional domains is still a challenge. Optimizing such functions by vanilla BO is extremely time-consuming. Alternative strategies for high-dimensional BO that are based on the idea of embedding the high-dimensional space to the one with low dimension are sensitive to the choice of the embedding dimension, which needs to be pre-specified. We develop a new computationally efficient high-dimensional BO method that exploits variable selection. Our method is able to automatically learn axis-aligned sub-spaces, i.e. spaces containing selected variables, without the demand of any pre-specified hyperparameters. We theoretically analyze the computational complexity of our algorithm. We empirically show the efficacy of our method on several synthetic and real problems.
\end{abstract}

\section{Introduction}

We study the problem of globally maximizing a black-box function $f(\mathbf{x})$ with an input domain $\mathcal{X}=[0,1]^{D}$, where the function has some special properties: (1) It is hard to calculate its first and second-order derivatives, therefore gradient-based optimization algorithms are not useful; (2) It is too strong to make additional assumptions on the function such as convexity; (3) It is expensive to evaluate the function, hence some classical global optimization algorithms such as evolutionary algorithms (EA) are not applicable.

Bayesian optimization (BO) is a popular global optimization method to solve the problem described above. It aims to obtain the input $\mathbf{x}^{*}$ that maximizes the function $f$ by sequentially acquiring queries that are likely to achieve the maximum and evaluating the function on these queries. BO has been successfully applied in many scenarios such as hyper-parameter tuning ~\citep{snoek2012practical,klein2017fast}, automated machine learning~\citep{nickson2014automated,yao2018taking}, reinforcement learning~\citep{brochu2010tutorial,marco2017virtual,wilson2014using}, robotics~\citep{calandra2016bayesian,berkenkamp2016bayesian}, and chemical design~\citep{griffiths2017constrained,negoescu2011knowledge}. However, most problems described above that have been solved by BO successfully have black-box functions with low-dimensional domains, typically with $D\leq 20$~\citep{frazier2018tutorial}. Scaling BO to high-dimensional black-box functions is challenging because of the following two reasons: (1) Due to the curse of dimensionality, the global optima is harder to find as $D$ increases; and (2) computationally, vanilla BO is extremely time consuming on functions with large $D$. As global optimization for high-dimensional black-box function has become a necessity in several scientific fields such as algorithm configuration~\citep{hutter2010automated}, computer vision~\citep{bergstra2013making} and biology~\citep{gonzalez2015bayesian}, developing new BO algorithms that can effectively optimize black-box functions with high dimensions is very important for practical applications. 

A large class of algorithms for high-dimensional BO is based on the assumption that the black-box function has an effective subspace with dimension $d_{e}\ll D$~\citep{djolonga2013high,wang2016bayesian,moriconi2019high,nayebi2019framework,letham2020re}. Therefore, these algorithms first embed the high-dimensional domain $\mathcal{X}$ to a space with the embedding dimension $d$ pre-specified by users, do vanilla BO in the embedding space to obtain the new query, and then project it back and evaluate the function $f$. These algorithms are time efficient since BO is done in a low-dimensional space. \citet{wang2016bayesian} proves that if $d\geq d_{e}$, then theoretically with probability $1$ the embedding space contains the global optimum. However, since $d_{e}$ is usually not known, it is difficult for users to set a suitable $d$. Previous work such as \citet{eriksson2021high} shows that different settings of $d$ will impact the performance of embedding-based algorithms, and there has been little work on how to choose $d$ heuristically. \citet{letham2020re} also points out that when projecting the optimal point in the embedding space back to the original space, it is not guaranteed that this projection point is inside $\mathcal{X}$, hence algorithms may fail to find an optimum within the input domain. 

We develop a new algorithm, called VS-BO (Variable Selection Bayesian Optimization), to solve issues mentioned above. Our method is based on the assumption that all the $D$ variables (elements) of the input $\mathbf{x}$ can be divided into two disjoint sets $\mathbf{x}=\{\mathbf{x}_{ipt},\mathbf{x}_{nipt}\}$: (1) $\mathbf{x}_{ipt}$, called important variables, are variables that have significant effects on the output value of $f$; (2) $\mathbf{x}_{nipt}$, called unimportant variables, are variables that have no or little effect on the output. Previous work such as \citet{hutter2014efficient} shows that the performance of many machine learning methods is strongly affected by only a small subset of hyperparameters, indicating the rationality of this assumption. We propose a robust strategy to identify $\mathbf{x}_{ipt}$, and do BO on the space of $\mathbf{x}_{ipt}$ to reduce time consumption. In particular, our method is able to learn the dimension of $\mathbf{x}_{ipt}$ automatically, hence there is no need to pre-specify the hyperparameter $d$ as embedding-based algorithms. Since the space of $\mathbf{x}_{ipt}$ is axis-aligned, issues caused by the space projection no longer exist in our method. We theoretically analyze the computational complexity of VS-BO, showing that our method can decrease the computational complexity of both steps of fitting the Gaussian Process (GP) and optimizing the acquisition function. We formalize the assumption that some variables of the input are important while others are unimportant and derive the regret bound of our method. Finally, we empirically show the good performance of VS-BO on several synthetic and real problems. 

The source code to reproduce the results from this study can be found at \url{https://github.com/Kingsford-Group/vsbo}. This work has been accepted in AutoML 2023, with camera-ready version \url{https://openreview.net/forum?id=QXKWSM0rFCK1}.

\section{Related work}

The basic framework of BO has two steps for each iteration: First, GP is used as the surrogate to model $f$ based on all the previous query-output pairs $\left(\mathbf{x}^{1:n},y^{1:n}\right)$: 
\begin{align*}
    y^{1:n}\sim \mathcal{N}\left(\mathbf{0},K(\mathbf{x}^{1:n},\Theta)+\sigma^{2}_{0}\mathbf{I}\right),
\end{align*}
Here $y^{1:n}=[y^{1},\dots , y^{n}]$ is a $n$-dimensional vector, $y^{i}=f(\mathbf{x}^{i})+\epsilon^{i}$ is the output of $f$ with random noise $\epsilon^{i}\sim\mathcal{N}(0,\sigma_{0}^{2})$, and $K(\mathbf{x}^{1:n},\Theta)$ is a $n\times n$ covariance matrix where its entry $K_{i,j}=k(\mathbf{x}^{i},\mathbf{x}^{j},\Theta)$ is the value of a kernel function $k$ in which $\mathbf{x}^{i}$ and $\mathbf{x}^{j}$ are the i-th and j-th queries respectively. $\Theta$ and $\sigma_{0}$ are parameters of GP that will be optimized each iteration, and $\mathbf{I}$ is the $n\times n$ identity matrix. A detailed description of GP and its applications can be found in \citet{williams2006gaussian}.

Given a new input $\mathbf{x}'$, we can compute the posterior distribution of $f(\mathbf{x}')$ from GP, which is again a Gaussian distribution with mean $\mu(\mathbf{x}'\mid \mathbf{x}^{1:n}, y^{1:n})$ and variance $\sigma^{2}(\mathbf{x}'\mid \mathbf{x}^{1:n})$ that have the following forms:
%\begin{align*}
%    &f(\mathbf{x}')\mid \mathbf{x}', \mathbf{x}^{1:n}, f(\mathbf{x}^{1:n}) \sim\mathcal{N}(\mu(\mathbf{x}'\mid \mathbf{x}^{1:n}, f(\mathbf{x}^{1:n})),\sigma^{2}(\mathbf{x}'\mid \mathbf{x}^{1:n}))
%\end{align*}
\begin{align*}
    & \mu(\mathbf{x}'\mid \mathbf{x}^{1:n}, y^{1:n})=\mathbf{k}(\mathbf{x}',\mathbf{x}^{1:n})[K(\mathbf{x}^{1:n},\Theta)+\sigma^{2}_{0}\mathbf{I}]^{-1}(y^{1:n})^{\top} \\
    &\sigma^{2}(\mathbf{x}'\mid \mathbf{x}^{1:n})=k(\mathbf{x}',\mathbf{x}',\Theta)-\mathbf{k}(\mathbf{x}',\mathbf{x}^{1:n})[K(\mathbf{x}^{1:n},\Theta)+\sigma^{2}_{0}\mathbf{I}]^{-1}\mathbf{k}(\mathbf{x}',\mathbf{x}^{1:n})^{\top}
\end{align*}
Here, $\mathbf{k}(\mathbf{x}',\mathbf{x}^{1:n})=[k(\mathbf{x}',\mathbf{x}^{1},\Theta),\dots , k(\mathbf{x}',\mathbf{x}^{n},\Theta)]$ is a $n$-dimensional vector. 

The second step of BO is to use $\mu$ and $\sigma$ to construct an acquisition function $acq$ and maximize it to get the new query $\mathbf{x}^{new}$, on which the function $f$ is evaluated to obtain the new pair $(\mathbf{x}^{new},y^{new})$:
\begin{align*}
    \mathbf{x}^{new} = \text{argmax}_{\mathbf{x}'\in \mathcal{X}}\; acq(\mu(\mathbf{x}'\mid \mathbf{x}^{1:n}, y^{1:n}),\sigma(\mathbf{x}'\mid \mathbf{x}^{1:n})).
\end{align*}

A wide variety of methods have been proposed that are related to high-dimensional BO, and most of them are based on some extra assumptions on intrinsic structures of the domain $\mathcal{X}$ or the function $f$. As mentioned in the previous section, a considerable body of algorithms is based on the assumption that the black-box function has an effective subspace with a significantly smaller dimension than $\mathcal{X}$. Among them, REMBO~\citep{wang2016bayesian} uses a randomly generated matrix as the projection operator to embed $\mathcal{X}$ to a low-dimensional subspace. SI-BO~\citep{djolonga2013high}, DSA~\citep{ulmasov2016bayesian} and MGPC-BO~\citep{moriconi2019high} propose different ways to learn the projection operator from data, of which the major shortcoming is that a large number of data points are required to make the learning process accurate. HeSBO~\citep{nayebi2019framework} uses a hashing-based method to do subspace embedding. Finally, ALEBO~\citep{letham2020re} aims to improve the performance of REMBO with several novel refinements. 

Another assumption is that the black-box function has an additive structure. \citet{kandasamy2015high} first develops a high-dimensional BO algorithm called Add-GP by adopting this assumption. They derive a simplified acquisition function and prove that the regret bound is linearly dependent on the dimension. Their framework is subsequently generalized by  \citet{li2016high,wang2017batched} and \citet{rolland2018high}.

As described in the previous section, our method is based on the assumption that some variables are more ``important" than others, which is similar to the axis-aligned subspace embedding. Several previous works propose different methods to choose axis-aligned subspaces in high-dimensional BO. \citet{li2016high} uses the idea of dropout, i.e, for each iteration of BO, a subset of variables are randomly chosen and optimized, while our work chooses variables that are important in place of the randomness. \citet{eriksson2021high} develops a method called SAASBO, which uses the idea of Bayesian inference. SAASBO defines a prior distribution for each parameter in the kernel function $k$, and for each iteration the parameters are sampled from posterior distributions and used in the step of optimizing the acquisition function. Since those priors restrict parameters to concentrate near zero, the method is able to learn a sparse axis-aligned subspaces (SAAS) during BO process. Similar to vanilla BO, the main drawback of SAASBO is that it is very time consuming. While traditionally it is assumed that the function $f$ is very expensive to evaluate so that the runtime of BO itself does not need to be considered, previous work such as \citet{ulmasov2016bayesian} points out that in some application scenarios the runtime of BO cannot be neglected. \citet{spagnol2019bayesian} proposes a similar framework of high-dimensional BO as us; they use Hilbert Schmidt Independence criterion (HSIC) to select variables, and use the chosen variables to do BO. However, they do not provide a comprehensive comparison with other high-dimensional BO methods: their method is only compared with the method in \citet{li2016high} on several synthetic functions. In addition, they do not provide any theoretical analysis. 

\begin{algorithm}
	\caption{VS-BO} 
	\begin{algorithmic}[1]
	    \State \textbf{Input}: $f(\mathbf{x})$, $\mathcal{X}=[0,1]^{D}$, $N_{init}$, $N$, $N_{vs}$
	    \State \textbf{Output}: Approximate maximizer $\mathbf{x}^{max}$
	    %Initialize  as $\emptyset$, initialize domain boundary $\mathcal{X}=[-1,1]^{M}$, maximal iteration budget $N$
	    \State Initialize the set of $\mathbf{x}_{ipt}$ to be all variables in $\mathbf{x}$, $\mathbf{x}_{ipt}=\mathbf{x}$, and $\mathbf{x}_{nipt}=\emptyset$
	    %\State Let $D_0=\emptyset$, $x_{ipt}=x$, $x_{nipt}=\emptyset$
	    \State Uniformly sample $N_{init}$ points $\mathbf{x}^{i}$ and evaluate $y^{i}=f(\mathbf{x}^{i})$, let $\mathcal{D}=\{(\mathbf{x}^{i},y^{i})\}_{i=1}^{N_{init}}$
	    \State Initialize the distribution $p(\mathbf{x}\mid \mathcal{D})$
	    %\For{$t=1,2,\ldots N_{rand}$}
	    %    \State $x^{t}\sim Unif(\mathcal{X})$
	    %    \State $y^{t}=f(x^{t})$
	    %    \State $D_{t}=D_{t-1}\cup{(x^t,y^t)}$
	    %\EndFor
		\For {$t=N_{init}+1,N_{init}+2,\ldots N_{init}+N$}
			\If {mod($t-N_{init}$, $N_{vs}$) = 0}
		        \State Variable selection to update $\mathbf{x}_{ipt}$ and let $\mathbf{x}_{nipt}=\mathbf{x}\setminus \mathbf{x}_{ipt}$ (Algorithm~\ref{alg:VS_detail})
		        \State Update $p(\mathbf{x}\mid \mathcal{D})$, then derive the conditional distribution $p(\mathbf{x}_{nipt}\mid \mathbf{x}_{ipt},\mathcal{D})$
		    \EndIf
		    \State Fit a GP to $\mathcal{D}_{ipt}:=\{(\mathbf{x}_{ipt}^{i},y^{i})\}_{i=1}^{t-1}$
		    \State Maximize the acquisition function to obtain $\mathbf{x}_{ipt}^{t}$.
		    %\State $x_{ipt}^{t+1}=\text{argmax}\; acq(x_{ipt})$
		    \State Sample $\mathbf{x}_{nipt}^{t}$ from $p(\mathbf{x}_{nipt}\mid \mathbf{x}_{ipt}^{t},\mathcal{D})$
		    %\State $x^{t+1}=\{x_{ipt}^{t+1},x_{nipt}^{t+1}\}$
		    \State Evaluate $y^{t}=f(\mathbf{x}^{t})+\epsilon^{t}=f(\{\mathbf{x}_{ipt}^{t},\mathbf{x}_{nipt}^{t}\})+\epsilon^{t}$ and update $\mathcal{D}=\mathcal{D}\cup \{(\mathbf{x}^{t},y^{t})\}$
		\EndFor
		\State\Return $\mathbf{x}^{max}$ which is equal to $\mathbf{x}^{i}$ with maximal $y^{i}$
	\end{algorithmic} 
	\label{alg:VSBO}
\end{algorithm}

\section{Framework of VS-BO}

Given the black-box function $f(\mathbf{x}): \mathcal{X}\to\mathbb{R}$ in the domain $\mathcal{X}=[0,1]^{D}$ with a large $D$, the goal of high-dimensional BO is to find the maximizer $\mathbf{x}^{*}=\text{argmax}_{\mathbf{x}\in \mathcal{X}}f(\mathbf{x})$ efficiently. As mentioned in the introduction, VS-BO is based on the assumption that all variables in $\mathbf{x}$ can be divided into important variables $\mathbf{x}_{ipt}$ and unimportant variables $\mathbf{x}_{nipt}$, and the algorithm uses different strategies to decide values of the variables from two different sets.

The high-level framework of VS-BO (Algorithm~\ref{alg:VSBO}) is similar to \citet{spagnol2019bayesian}. For every $N_{vs}$ iterations VS-BO will update $\mathbf{x}_{ipt}$ and $\mathbf{x}_{nipt}$ (line 8 in Algorithm~\ref{alg:VSBO}), and for every BO iteration $t$ only variables in $\mathbf{x}_{ipt}$ are used to fit GP (line 11 in Algorithm~\ref{alg:VSBO} ), and the new query of important variables $\mathbf{x}_{ipt}^{t}$ is obtained by maximising the acquisition function (line 12 in Algorithm~\ref{alg:VSBO}). Unlike \citet{spagnol2019bayesian}, VS-BO learns a conditional distribution $p(\mathbf{x}_{nipt}\mid \mathbf{x}_{ipt},\mathcal{D})$ from the existing query-output pairs $\mathcal{D}$ (line 5, 9 in Algorithm~\ref{alg:VSBO}). This distribution is used for choosing the value of $\mathbf{x}_{nipt}$ to make $f(\mathbf{x})$ large when $\mathbf{x}_{ipt}$ is fixed. Hence, once $\mathbf{x}_{ipt}^{t}$ is obtained, the algorithm samples $\mathbf{x}_{nipt}^{t}$ from $p(\mathbf{x}_{nipt}\mid \mathbf{x}_{ipt}^{t},\mathcal{D})$ (line 13 in Algorithm~\ref{alg:VSBO}), concatenates it with $\mathbf{x}_{ipt}^{t}$ and evaluates $f(\{\mathbf{x}_{ipt}^{t},\mathbf{x}_{nipt}^{t}\})$. 

Compared to \citet{spagnol2019bayesian}, our method is new on the following three aspects: First, we propose a new variable selection method that takes full advantage of the information in the fitted GP model, and there is no hyperparameter that needs to be pre-specified in this method; Second, we develop a new mechanism, called VS-momentum, to improve the robustness of variable selection; Finally, we integrate an evolutionary algorithm into the framework of BO to make the sampling of unimportant variables more precise. The following subsections introduce these three points in detail. 

%slfdasfasd aasdfasdf afasfsad fasdfasasf asdfasf asdfasdf afasd fasdfasd fas fasdf asdf sdf sadfas dfasd fsd fasdf asdf asf asfds fdf dfd fdfd fgasg adgdas gadsg dg adsg adsgasd gsdf asfa sdgsd gas gasg adsgads gas gadsg asg asdga sdgas gasdg sags adgas gadsg asg sdagsdr  afdf adsf adsfdas gads gads gadg adsg adsg dsgds gds gsd gsdg dsg adsg adsg adsg adsg dsgsd gsd gds gds gsd gdsa gasdg sdg dsg sadgds gas gdsa gdsa gasdg adsg dsg sadg dg asdg dg asdgasd gasd gads gads gdsag dsg adsg adsg adsg dsg asdgsd gds gds gads gdsg sadg dsgsd gsa gadg dsgag asfd f fs fds faf sdf dsfas fds fds g gasgds gdas ga

\begin{algorithm}
	\caption{Variable Selection (line 8 in Algorithm~\ref{alg:VSBO})} 
	\begin{algorithmic}[1]
	    \State \textbf{Input}: $\mathcal{D}=\{(\mathbf{x}^{i},y^{i})\}_{i=1}^{t}$
	    \State \textbf{Output}: Set of important variables $\mathbf{x}_{ipt}$
	    \State Fit a GP to $\mathcal{D}$ and calculate important scores of variables $IS$ where $IS[i]$ is the important score of the i-th variable 
	    %\State Fit GP using all variables: $Y=(y^{1:t})\sim \mathcal{N}(\mathbf{0},K(x^{1:t},\Theta)+\sigma^{2}_{t}I)$, maximize the marginal likelihood and record the final loss $L_0$.
	    %\State Calculate important scores of variables, $IS$, $IS[i]$ is the important score of the i-th variable. 
	    \State Sort variables according to their important scores, $[\mathbf{x}_{s(1)},\dots, \mathbf{x}_{s(D)}]$, from the most important to the least
	    %\State Fit GP using the most important variables: $Y=(y^{1:t})\sim \mathcal{N}(\mathbf{0},K(x^{1:t}_{s(1)},\Theta)+\sigma^{2}_{t}I)$, maximize the marginal likelihood and record the final loss $L_1$.
	    \For{$m=1,2,\ldots D$} \Comment{Stepwise forward selection}
	        \State Fit a GP to $\mathcal{D}_{m}:=\{(\mathbf{x}_{s(1):s(m)}^{i},y^{i})\}_{i=1}^{t-1}$ where $\mathbf{x}_{s(1):s(m)}^{i}$ is the $i$-th input with only the first $m$ important variables, let $L_m$ to be the value of final negative marginal log likelihood
	        \If{$m<3$}
	            \State \textbf{continue}
	        %\State Fit GP using the first $m$ important variables: $Y=(y^{1:t})\sim \mathcal{N}(\mathbf{0},K(x^{1:t}_{s(1):s(m)},\Theta)+\sigma^{2}_{t}I)$, maximize the marginal likelihood and record the final loss $L_m$.
	        \ElsIf {$L_{m-1}-L_{m}\leq 0$ or $L_{m-1}-L_{m}<\frac{L_{m-2}-L_{m-1}}{10}$}
	            \State \textbf{break}
	        \EndIf
	    \EndFor
	    \State\Return
	    $\mathbf{x}_{ipt}=\{\mathbf{x}_{s(1)},\dots, \mathbf{x}_{s(m-1)}\}$
	\end{algorithmic} 
	\label{alg:VS_detail}
\end{algorithm}

\subsection{Variable selection}

The variable selection step in VS-BO (Algorithm~\ref{alg:VS_detail}) can be further separated into two substeps: (1) calculate the importance score ($IS$) of each variable (line 3 in Algorithm~\ref{alg:VS_detail}), and (2) do the stepwise-forward variable selection~\citep{derksen1992backward} according to the importance scores.

For step one, we develop a gradient-based $IS$ calculation method, called Grad-IS, inspired by \citet{paananen2019variable}. Intuitively, if the partial derivative of the function $f$ with respect to one variable is large on average, then the variable ought to be important. Since the derivative of $f$ is unknown, VS-BO instead estimates the expectation of the gradient of posterior mean from a fitted GP model, normalized by the posterior standard deviation: 
\begin{align*}
    IS &=\mathbb{E}_{\mathbf{x}\sim Unif(\mathcal{X})}\left[\frac{\nabla_{\mathbf{x}}\mathbb{E}_{p(f(\mathbf{x})\mid \mathbf{x}, \mathcal{D})}\left[f(\mathbf{x}) \right]}{\sqrt{Var_{p(f(\mathbf{x})\mid \mathbf{x}, \mathcal{D})}\left[f(\mathbf{x}) \right]}} \right] = \mathbb{E}_{\mathbf{x}\sim Unif(\mathcal{X})}\left[\frac{\nabla_{\mathbf{x}}\mu(\mathbf{x}\mid \mathcal{D})}{\sigma(\mathbf{x}\mid \mathcal{D})} \right] \\
    &\approx\frac{1}{N_{is}}\sum_{k=1}^{N_{is}}\frac{\nabla_{\mathbf{x}}\mu(\mathbf{x}^{k}\mid \mathcal{D})}{\sigma(\mathbf{x}^{k}\mid \mathcal{D})}\quad \mathbf{x}^{k} \stackrel{i.i.d}{\sim}Unif(\mathcal{X}).
\end{align*}
Here, both $\nabla_{\mathbf{x}}\mu(\cdot\mid \mathcal{D})$ and $\sigma(\cdot\mid \mathcal{D})$ have explicit forms. Both the Grad-IS and Kullback-Leibler Divergence (KLD)-based methods in \citet{paananen2019variable} are estimations of $\mathbb{E}_{\mathbf{x}\sim Unif(\mathcal{X})}\left[\frac{\nabla_{\mathbf{x}}\mathbb{E}_{p(f(\mathbf{x})\mid \mathbf{x}, \mathcal{D})}\left[f(\mathbf{x}) \right]}{\sqrt{Var_{p(f(\mathbf{x})\mid \mathbf{x}, \mathcal{D})}\left[f(\mathbf{x}) \right]}} \right]$. Since the KLD method only calculates approximate derivatives around the chosen points in $\mathcal{D}$ that are always unevenly distributed, it is a biased estimator, while our importance score estimation is unbiased. 

Each time the algorithm fits GP to the existing query-output pairs, the marginal log likelihood (MLL) of GP is maximized by updating parameters $\Theta$  and $\sigma_{0}$. VS-BO takes negative MLL as the loss and uses its value as the stopping criteria of the stepwise-forward selection. More specifically, VS-BO sequentially selects variables according to the important score, and when a new variable is added, the algorithm will fit GP again by only using those chosen variables and records a new final loss (line 6 in Algorithm~\ref{alg:VS_detail}). If the new loss is nearly identical to the previous loss, the loss of fitted GP when the new variable is not included, then the selection step stops (line 9 in Algorithm~\ref{alg:VS_detail}) and all those already chosen variables are important variables. 

Consider the squared exponential kernel, a common kernel choice for GP, which is given by
\begin{align*}
    k(\mathbf{x},\mathbf{x}',\Theta=\{\rho^{2}_{1:D},\alpha_{0}^{2}\}) = \alpha_{0}^{2}\exp\left( -\frac{1}{2}\sum_{i=1}^{D}\rho_{i}^{2}(\mathbf{x}_{i}-\mathbf{x}'_{i})^2 \right),
\end{align*}

where $\rho_{i}^{2}$ is the inverse squared length scale of the $i$-th variable. On the one hand, when only a small subset of variables in $\mathbf{x}$ are important, the variable selection is similar to adding a $L_{0}$ regularization for GP fitting step.  Let $\mathbf{\rho}=[\rho^{2}_{1},\dots , \rho^{2}_{D}]$, the variable selection step chooses a subset of variables and specifies $\rho_{i}^{2}=0$ when $i$-th variable is in $\mathbf{x}_{nipt}$, leading $\norm{\mathbf{\rho}}_{0}$ to be small. Therefore, fitting GP by only using variables in $\mathbf{x}_{ipt}$ is similar to learning the kernel function with sparse parameters. On the other hand, when in the worst case every variable is equally important, $\mathbf{x}_{ipt}$ is likely to contain nearly all the variables in $\mathbf{x}$, and in that case VS-BO degenerates to vanilla BO.

\subsection{Momentum mechanism in variable selection}

The idea of VS-momentum is to some extent similar to momentum in the stochastic gradient descent~\citep{loizou2017momentum}. Intuitively, queries obtained after one variable selection step can give extra information on the accuracy of this variable selection. If empirically a new maximizer is found, then this variable selection step is likely to have found real important variables, hence most of these variables should be kept at the next variable selection step. Otherwise, most should be removed and new variables need to be added. 

\begin{figure}
  \centering
  %\fbox{\rule[-.5cm]{0cm}{4cm} \rule[-.5cm]{4cm}{0cm}}
  \includegraphics[width=0.90\textwidth]{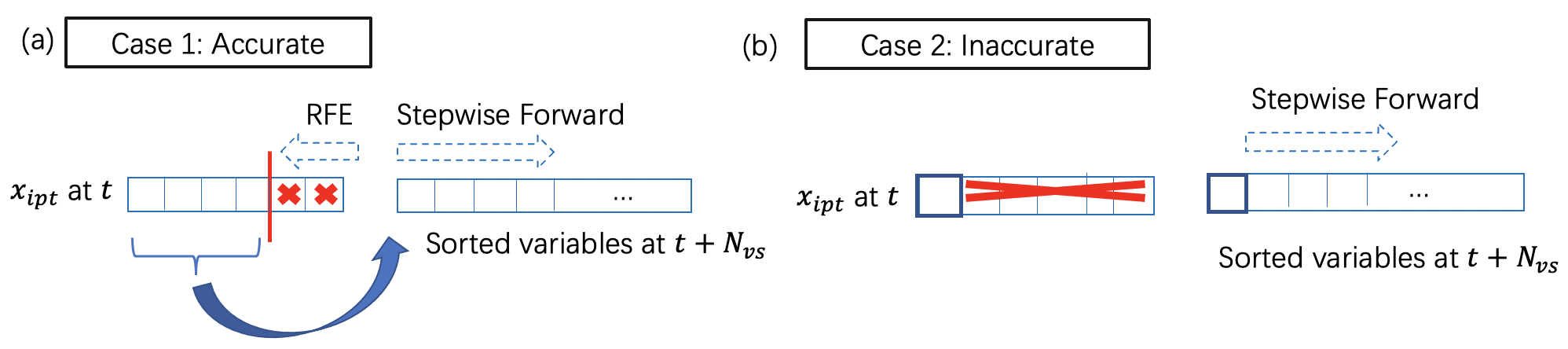}
  \caption{Momentum mechanism in VS-BO. (a) Accurate case, RFE is first used to remove redundant variables, and then new variables are added. (b) Inaccurate case, most variables are removed except those that are considered very important in both variable selection steps (blue box). New variables are then added.}
\label{fig:momentum}
\end{figure}

More specifically, we say that the variable selection at iteration $t+N_{vs}$ is in an accurate case when $\max_{k\in \{t+1,\dots, t+N_{vs}\}}y^{k}>\max_{k\in \{1,\dots, t\}}y^{k}$, otherwise it is in an inaccurate case. In the accurate case, VS-BO first uses recursive feature elimination (RFE) based algorithm to remove redundant variables in $\mathbf{x}_{ipt}$ that is selected at $t$, then it adds new variables into the remaining only if the loss decreases evidently (Figure~\ref{fig:momentum}a). In the inaccurate case, variables selected at $t$ will not be considered at $t+N_{vs}$ unless they still obtain very high important scores at $t+N_{vs}$ (marked by the blue box in Figure~\ref{fig:momentum}b). New variables are added via stepwise-forward algorithm. The details of variable selection with momentum mechanism are described in section A of the appendix. 

\subsection{Sampling for unimportant variables}

We propose a method based on Covariance Matrix Adaptation Evolution Strategy (CMA-ES) to obtain the new value of unimportant variables for each iteration. CMA-ES is an evolutionary algorithm for numerically optimizing a function. For each generation $k$, the algorithm samples new offsprings from a multivariate Gaussian distribution $\mathcal{N}\left(m^{(k-1)},(\sigma^{(k-1)})^2\right)$ and updates $m^{(k-1)}$ and $(\sigma^{(k-1)})^2$ based on these new samples and their corresponding function values. Details of this algorithm can be seen in \citet{hansen2016cma}.

Using the same approach as CMA-ES, VS-BO uses the initialized data $\{(\mathbf{x}^{i},y^{i})\}_{i=1}^{N_{init}}$ to initialize the multivariate Gaussian distribution $p(\mathbf{x}\mid \mathcal{D})$ (line 5 in Algorithm~\ref{alg:VSBO}), and for every $N_{vs}$ iterations, it updates the distribution based on new query-output pairs (line 9 in Algorithm~\ref{alg:VSBO}). Because of the property of Gaussian distribution, the conditional distribution $p(\mathbf{x}_{nipt}\mid \mathbf{x}_{ipt},\mathcal{D})$ is easily derived which is also a multivariate Gaussian distribution. Therefore, $\mathbf{x}_{nipt}^{t}$ can be sampled from the Gaussian distribution $p(\mathbf{x}_{nipt}\mid \mathbf{x}_{ipt}^{t},\mathcal{D})$ (line 13 in Algorithm~\ref{alg:VSBO}) when $\mathbf{x}_{ipt}^{t}$ is obtained. 

Compared to BO, it is much faster to update the evolutionary algorithm and obtain new queries, although these queries are less precise than those from BO. VS-BO takes advantage of the strength of these two methods by using them on different variables. Important variables are crucial to the function value, therefore VS-BO uses the framework of BO on them to obtain precise queries. Unimportant variables do not effect the function value too much so there is no need to spend large time budget to search for extremely precise queries. Hence,  they are determined by CMA-ES to reduce runtime. In addition, when the variable selection step is inaccurate, VS-BO degenerates to an algorithm that is similar to CMA-ES rather than random sampling, therefore this sampling strategy may help improve the robustness of the performance of the whole algorithm. 

\section{Computational complexity analysis}

From the theoretical perspective, we prove that running BO by only using those important variables is able to decrease the runtime of both the step of fitting the GP and maximizing the acquisition function. Specifically, we have the following proposition: 

\begin{prop}\label{prop:complexity}
    Suppose the cardinality of $\mathbf{x}_{ipt}$ is $p$ and the Quasi-Newton method (QN) is used for both fitting the GP and maximizing the acquisition function. Under the choice of commonly used kernel functions and acquisition functions, if only variables in $\mathbf{x}_{ipt}$ is used, then the complexity of each step of QN is $\mathcal{O}(p^2+pn^2+n^3)$ for fitting the GP and $\mathcal{O}(p^2+pn+n^2)$ for maximizing the acquisition function, where $n$ is the number of queries that are already obtained. 
\end{prop}

The proof is in section B of the appendix. Note that the method for fitting the GP and maximizing the acquisition function under the framework of BoTorch is limited-memory BFGS, which is indeed a QN method. Since the complexity is related to the quadratic of $p$, selecting a small subset of variables (so that $p$ is small) can decrease the runtime of BO. Figure~\ref{fig:runtime_compare_branin} empirically shows that compared to vanilla BO, VS-BO can both reduce the runtime of fitting a GP and optimizing the acquisition function, especially when $n$ is not small. 

\section{Regret bound analysis}

Let $\mathbf{x}^{*}$ be one of the maximal points of $f(\mathbf{x})$. To quantify the efficacy of the optimization algorithm, we are interested in the cumulative regret $R_{N}$, defined as: $R_{N}=\sum_{t=1}^{N}\left[f(\mathbf{x}^{*})-f(\mathbf{x}^{t})\right]$ where $\mathbf{x}^{t}$ is the query at iteration $t$. Intuitively, the algorithm is better when $R_{N}$ is small, and a desirable property is to have no regret: $\lim_{N\to\infty}R_{N}/N=0$. Here, we provide an upper bound of the cumulative regret for a simplified VS-BO algorithm, called VS-GP-UCB (Algorithm~\ref{alg:VS_GP_UCB} in the appendix). Similar to \citet{srinivas2009gaussian}, for proving the regret bound we need the smoothness assumption of the kernel function. In addition, we have the extra assumption that the $D-d$ variables in $\mathbf{x}$ are unimportant (for convenience we index unimportant variables from $d+1$ to $D$ without loss of generality), meaning the absolute values of partial derivatives of $f$ on those $D-d$ variables are in general smaller than those on important variables. Formally, we have the following assumption:

%In VS-GP-UCB, the last $D-d$ variables in $\mathbf{x}$ that are fixed before BO iterations (line 4 in Algorithm~\ref{alg:VS_GP_UCB}), denoted as $\mathbf{x}^{0}_{[d+1:D]}$, then the first $d$ variables are queried by maximising upper confidence bound (UCB)~\citep{auer2002using} with fixed unimportant variables (line 6 in Algorithm~\ref{alg:VS_GP_UCB}). Similar to \citet{srinivas2009gaussian}, we need the smoothness assumption of the kernel function, in addition, we have the extra assumptions that the $D-d$ variables are unimportant (for convenience we index unimportant variables from $d+1$ to $D$ without loss of generality):

\begin{assumption} \label{assump:derivative}
    Let $\mathcal{X}\subset[0,1]^{D}$ be compact and convex, $D\in \mathbb{N}$, and $f$ be a sample path of a GP with mean zero and the kernel function $k$, which satisfies the following high probability bound on the derivatives of $f$ for some constants $a,b>0$, $1>\alpha\geq 0$:
    \begin{align*}
    P(\sup_{\mathbf{x}\in \mathcal{X}}\abs{\frac{\partial f}{\partial  \mathbf{x}_j}}>L) \leq a\exp\left(-\left(\frac{L}{b}\right)^2\right), j=1,\dots , d
\end{align*}
And:
\begin{align*}
    P(\sup_{\mathbf{x}\in \mathcal{X}}\abs{\frac{\partial f}{\partial \mathbf{x}_j}}> L) \leq a\exp\left(-\left(\frac{L}{\alpha b}\right)^2\right), j=d+1,\dots , D,
\end{align*}
\end{assumption}

In VS-GP-UCB, the values of unimportant variables are fixed in advance (line 4 in Algorithm~\ref{alg:VS_GP_UCB}), denoted as $\mathbf{x}^{0}_{[d+1:D]}$, and the important variables are queried at each iteration by maximizing the acquisition function upper confidence bound (UCB)~\citep{auer2002using} with those fixed unimportant variables (line 6 in Algorithm~\ref{alg:VS_GP_UCB}). We have the following regret bound theorem of VS-GP-UCB:
\begin{theorem} \label{thm:regret}
    Let $\mathcal{X}\subset[0,1]^{D}$ be compact and convex, suppose Assumption \ref{assump:derivative} is satisfied, pick $\delta\in (0,1)$, and define 
    \begin{align*}
        \beta_{t}= 2\log\frac{8\pi^{2}t^{2}}{3\delta} + 2(D-d)\log\left(\alpha Dt^{2}b\sqrt{\log\left(\frac{8Da}{\delta}\right)}+1\right) + 2d\log\left( Dt^{2}b\sqrt{\log\left(\frac{8Da}{\delta}\right)}\right).
    \end{align*}
    Running the VS-GP-UCB, with probability $\geq 1-\delta$, we have:
    \begin{align*}
        \frac{R_{N}}{N} = \frac{\sum_{t=1}^{N}r_{t}}{N}\leq 2\sqrt{C_1 \frac{\beta_{N}\gamma_{N}}{N}} + \frac{\pi^{2}}{3N} + \alpha b\sqrt{\log\left(\frac{8Da}{\delta}\right)}(D-d),
    \end{align*}
\end{theorem}
Here, $\gamma_{N}:=\max_{A\subset \mathcal{X}: |A|=N}\mathbf{I}(\mathbf{y}_{A};\mathbf{f}_{A})$ is the maximum information gain with a finite set of sampling points $A$, $\mathbf{f}_{A}=[f(\mathbf{x})]_{\mathbf{x}\in A}$, $\mathbf{y}_{A}=\mathbf{f}_{A} + \epsilon_{A}$, and $C_1=\frac{8}{\log\left(1+\sigma_{0}^{-2}\right)}$.

%where $C_1=\frac{8}{\log\left(1+\sigma_{0}^{-2}\right)}$, and $\gamma_{N}:=\max_{A\subset \mathcal{X}: |A|=N}\mathbf{I}(\mathbf{y}_{A};\mathbf{f}_{A})$ is the maximum information gain with a finite set of sampling points $A$, $\mathbf{f}_{A}=[f(\mathbf{x})]_{\mathbf{x}\in A}$, $\mathbf{y}_{A}=\mathbf{f}_{A} + \epsilon_{A}$.

The proof of Theorem \ref{thm:regret} is in section C of the appendix. \citet{srinivas2009gaussian} upper bounded the maximum information gain for some commonly used kernel functions, for example they prove that by using SE kernel with the same length scales ($\rho_{i}=\rho_{0}$ for all $i$), $\gamma_{N}=\mathcal{O}\left((\log N)^{D+1}\right)$ so that $\lim_{N\to\infty}(\beta_{N}\gamma_{N})/N=0$. However, every variable is equally important in the SE kernel with the same length scales, which does not obey Assumption \ref{assump:derivative}. We hypothesize that by using SE kernel of which the length scales are different such that Assumption \ref{assump:derivative} is satisfied, the statement that $\lim_{N\to\infty}(\beta_{N}\gamma_{N})/N=0$ is also correct, although we do not have proof here. 

\citet{li2016high} also derives a regret bound for its dropout algorithm (Lemma 5 in \citet{li2016high}). Compared to the regret bound in \citet{srinivas2009gaussian} (Theorem 2 in \citet{srinivas2009gaussian}), both \citet{li2016high} and our work have an additional residual in the bound, while ours contains a small coefficient $\alpha$. In the case when $\alpha\to 0$, the bound in Theorem \ref{thm:regret} is the same as that in theorem 2 of \citet{srinivas2009gaussian} and there is no regret. These results show the necessity of the variable selection since it can help decrease the value of $\alpha$. In addition, compared to fixing unimportant variables in VS-GP-UCB, sampling from the CMA-ES posterior may further decrease the residual value. 

\section{Experiments}

We compare VS-BO to a broad selection of existing methods: vanilla BO, REMBO and its variant REMBO Interleave, Dragonfly, HeSBO and ALEBO. The details of implementations of these methods as well as hyperparameter settings  are described in section D of the appendix. 

\begin{figure}[!ht]
  \centering
  %\fbox{\rule[-.5cm]{0cm}{4cm} \rule[-.5cm]{4cm}{0cm}}
  \includegraphics[width=0.99\textwidth]{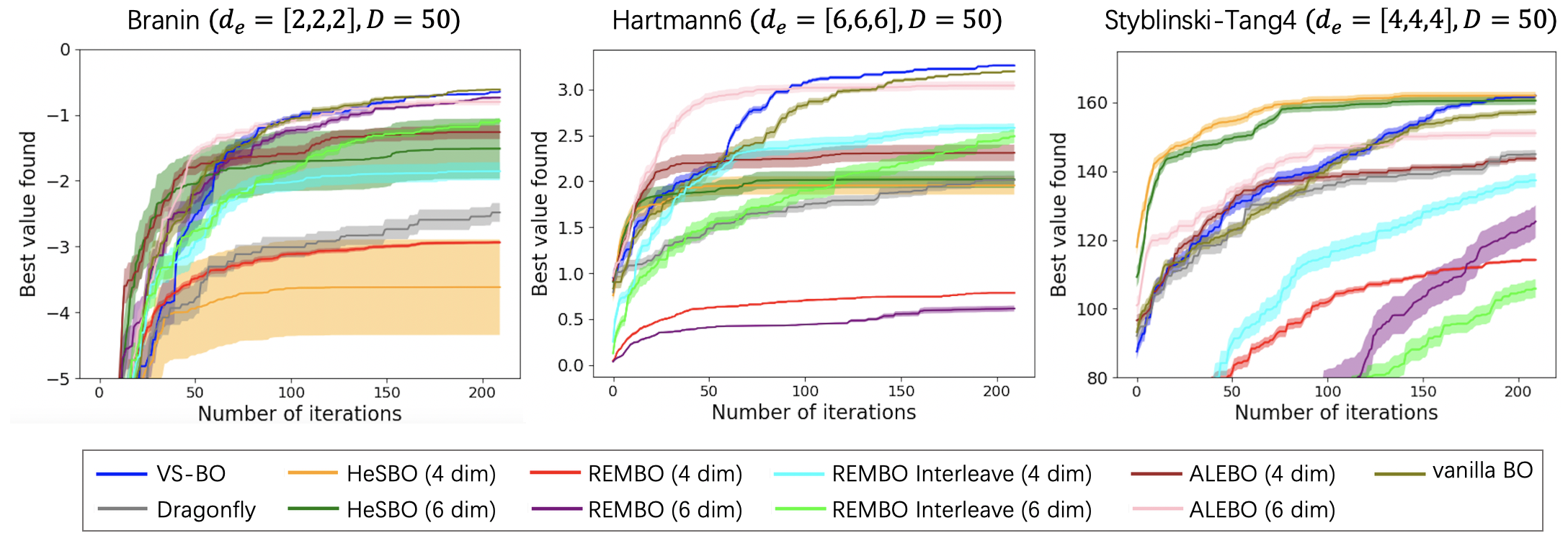}
  \caption{Performance of BO methods on Branin, Hartmann6 and Styblinski-Tang4 test functions. For each test function, we do 20 independent runs for each method. We plot the mean and 1/8 standard deviation of the best maximum value found by iterations.}
\label{fig:sythentic_result_iter}
\end{figure}

\subsection{Synthetic problems}

We use the Branin ($d_{e}=2$), Hartmann6 ($d_{e}=6$) and Styblinski-Tang4 ($d_{e}=4$) functions as test functions. Previous high-dimensional BO work extends these functions to high dimension by adding unrelated variables, while in our work we present a harder test setting that has not been tried before by adding both unrelated and unimportant (but not totally unrelated) variables. For example, in the Hartmann6 case with the standard Hartmann6 function $f_{Hartmann6}(\mathbf{x}_{[1:6]})$ we first construct a new function $F_{hm6}(\mathbf{x})$ by adding variables with different importance, $F_{hm6}(\mathbf{x})=f_{Hartmann6}(\mathbf{x}_{[1:6]})+0.1f_{Hartmann6}(\mathbf{x}_{[7:12]})+0.01f_{Hartmann6}(\mathbf{x}_{[13:18]})$, and we further extend it to $D=50$ by adding unrelated variables; see section D for full details. The dimension of effective subspace of $F_{hm6}$ is $18$, while the dimension of important variables is only $6$. We hope that VS-BO can find those important variables successfully. For each embedding-based methods we evalualte both $d=4$ and $d=6$. 

Figures~\ref{fig:sythentic_result_iter} and ~\ref{fig:sythentic_result_time} show performance of VS-BO as well as other BO methods on these three synthetic functions. When the iteration budget is fixed (Figure~\ref{fig:sythentic_result_iter}), the best value in average found by VS-BO after $200$ iterations is the largest or slightly smaller than the largest in all three cases. When the wall clock time or CPU time budget for BO is fixed (Figure~\ref{fig:sythentic_result_time}), results show that VS-BO can find a large function value with high computational efficiency. Figure~\ref{fig:sythentic_f_chosen} shows that VS-BO can accurately find all the real important variables and meanwhile control false positives. We also test VS-BO on the function that has a non-axis-aligned subspace, and results in Figure~\ref{fig:rotation_compare_iter} show that VS-BO also performs well. Vanilla BO under the framework of BoTorch can also achieve good performance for the fixed iteration budget, however, it is very computationally inefficient. For embedding-based methods, the results reflect some of their shortcomings. First, the performance of these methods are more variable than VS-BO; for example, HeSBO with $d=6$ performs very well in the Styblinski-Tang4 case but not in the others; Second, embedding-based methods are sensitive to the choice of the embedding dimension $d$, they perform especially bad when $d$ is smaller than the dimension of important variables (see results of the Hartmann6 case) and may still perform not well even when $d$ is larger (such as ALEBO with $d=6$ in the Styblinski-Tang4 case), while VS-BO can automatically learn the dimension. One advantage of embedding-based methods is that they may have a better performance than VS-BO within a very limited iteration budget (for example 50 iterations), which is expected since a number of data points are needed for VS-BO to make the variable selection accurate. 
%REMBO with $d=6$ has a great performance on Branin case but very bad performances on the other, ALEBO with $d=6$ performs well on both Branin and Hartmann6 cases but not on Styblinski-Tang4 case

\subsection{Real-world problems}

We compare VS-BO with other methods on two real-world problems. First, VS-BO is tested on the rover trajectory optimization problem presented in \citet{wang2017batched}, a problem with a $60$-dimensional input domain. Second, it is tested on the vehicle design problem MOPTA08~\citep{jones2008large}, a problem with $124$ dimensions. On these two problems, we evaluate both $d=6$ and $d=10$ for each embedding-based method, except we omit ALEBO with $d=10$ since it is very time consuming. The detailed settings of these two problems are described in section D of the appendix. 

\begin{figure}
  \centering
  %\fbox{\rule[-.5cm]{0cm}{4cm} \rule[-.5cm]{4cm}{0cm}}
  \includegraphics[width=0.88\textwidth]{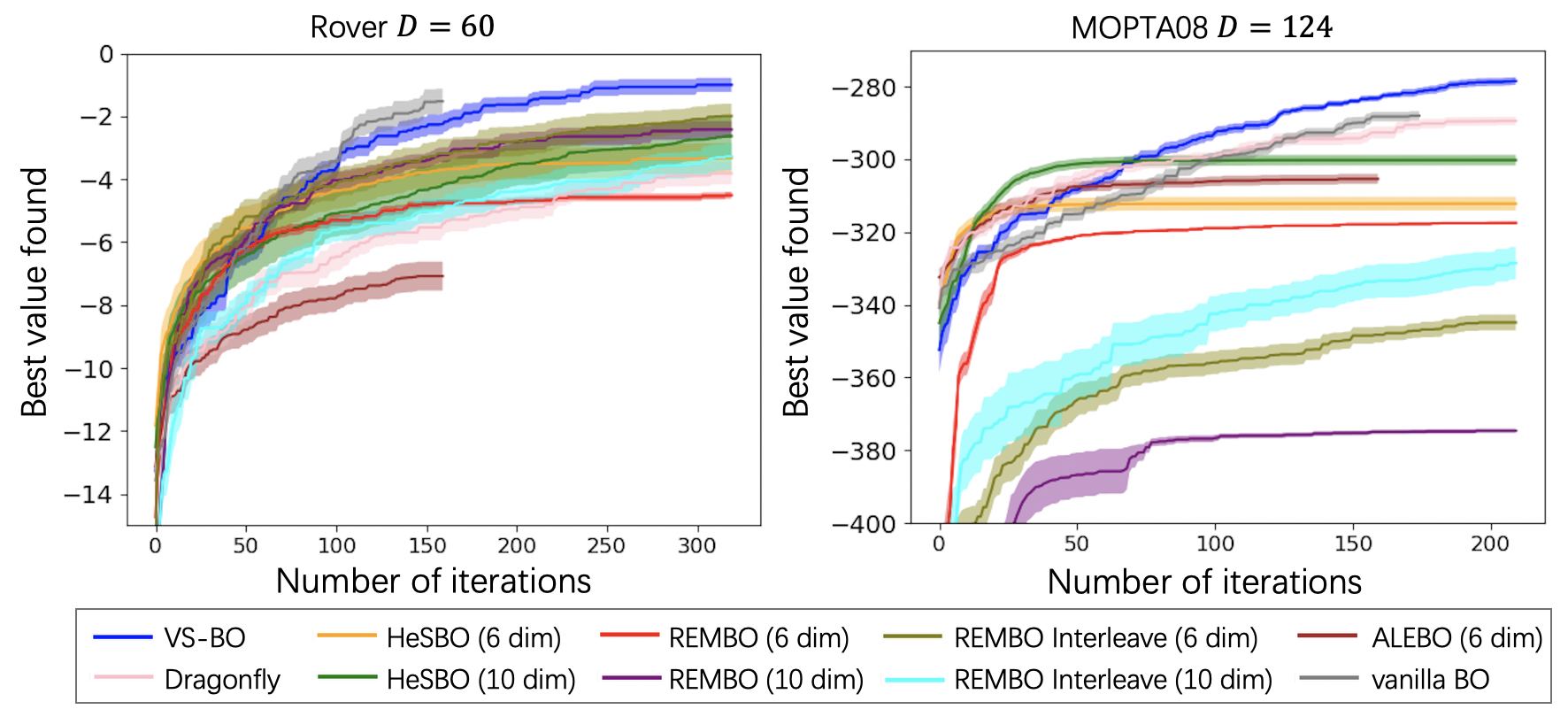}
  \caption{Performance of BO methods on the rover trajectory and MOPTA08 problems. We do 20 independent runs on the rover trajectory problem and 15 on the MOPTA08 problem. We plot the mean and 1/4 standard deviation of the best maximum value found by iterations. Curves of vanilla BO and ALEBO with $d=6$ do not reach the maximum iteration since they are time consuming and cannot run the maximum within the wall clock time budget (3600 seconds for the rover trajectory problem for each run and 4800 seconds for the MOPTA08 problem).}
\label{fig:real_result_iter}
\end{figure}

Figures~\ref{fig:real_result_iter} and ~\ref{fig:real_result_time} show the performance of VS-BO and other BO methods on these two problems. When the iteration budget is fixed (Figure~\ref{fig:real_result_iter}), VS-BO and vanilla BO have a better performance than other methods on both problems. When the wall clock or CPU time is fixed (Figure~\ref{fig:real_result_time}),  Dragonfly and VS-BO reach the best performance on the rover trajectory problem, and Dragonfly performs the best on MOPTA08 problem while VS-BO has the second best performance. Vanilla BO is computationally inefficient so it does not have good performance with the fixed runtime. The left column of Figure~\ref{fig:real_F_permutation} shows the frequency of being chosen as important for each variable when VS-BO is used. Since there is no ground truth of important variables in real-world cases, we use a sampling experiment to test whether those more frequently-chosen variables are more important. Specifically, we sample the first $5$ variables that have been chosen most frequently by a Sobol sequence and fix the values of other variables with the values in the best query we have found (the query having the highest function value). We then calculate the function values of this set of samples. Likewise, we also sample the first $5$ variables that have been chosen least frequently and evaluate the functions. The right column of Figure~\ref{fig:real_F_permutation} shows that the variance of function values from the first set of samples is significantly higher than that from the second, especially on the MOPTA08 problem, indicating that those frequently selected variables indeed have more significant effect on the function value.  

%\subsection{mopta}

\section{Conclusion}

%\textcolor{red}{1.proof, 2. curse of dimensionality, 3. rembo interleave}

We propose a new method, VS-BO, for high-dimensional BO that is based on the assumption that variables of the input can be divided to two categories: important and unimportant. Our method can assign variables into these two categories with no need for pre-specifying any crucial hyper-parameter and use different strategies to decide values of the variables in different categories to reduce runtime. The good performance of our method on synthetic and real-world problems further verify the rationality of the assumption. We show the computational efficiency of our method both theoretically and empirically. In addition, information from the variable selection improves the interpretability of BO model: VS-BO can find important variables so that it can help increase our understanding of the black-box function. We also notice that in practice vanilla BO under the framework of BoTorch usually has a good performance if the runtime of BO does not need to be considered, especially when the dimension is not too large ($D<100$). However, this method is usually not considered as a baseline to compare with in previous high-dimensional BO work. 

We also find some limitations of our method when running experiments. First, when the dimension of the input increases, it becomes harder to do variable selection accurately. Therefore, embedding-based methods are still the first choice when the input of a function has thousands of dimensions. It might be interesting to develop new algorithms that can do variable selection robustly even when the dimension is extremely large. Further, Grad-IS might be invalid when variables are discrete or categorical, therefore new methods for calculating the importance score of these kinds of variables are needed. These are several directions for future improvements of VS-BO. It is also interesting to do further theoretical study such as investigating the bounds on the maximum information gain of kernels that satisfy Assumption \ref{assump:derivative}.

\bibliographystyle{plainnat}

\bibliography{citations}

\begin{thebibliography}{40}
\providecommand{\natexlab}[1]{#1}
\providecommand{\url}[1]{\texttt{#1}}
\expandafter\ifx\csname urlstyle\endcsname\relax
  \providecommand{\doi}[1]{doi: #1}\else
  \providecommand{\doi}{doi: \begingroup \urlstyle{rm}\Url}\fi

\bibitem[Auer(2002)]{auer2002using}
Peter Auer.
\newblock Using confidence bounds for exploitation-exploration trade-offs.
\newblock \emph{Journal of Machine Learning Research}, 3\penalty0
  (Nov):\penalty0 397--422, 2002.

\bibitem[Bergstra et~al.(2013)Bergstra, Yamins, and Cox]{bergstra2013making}
James Bergstra, Daniel Yamins, and David Cox.
\newblock Making a science of model search: Hyperparameter optimization in
  hundreds of dimensions for vision architectures.
\newblock In \emph{International conference on machine learning}, pages
  115--123, 2013.

\bibitem[Berkenkamp et~al.(2016)Berkenkamp, Krause, and
  Schoellig]{berkenkamp2016bayesian}
Felix Berkenkamp, Andreas Krause, and Angela~P Schoellig.
\newblock Bayesian optimization with safety constraints: safe and automatic
  parameter tuning in robotics.
\newblock \emph{arXiv preprint arXiv:1602.04450}, 2016.

\bibitem[Brochu et~al.(2010)Brochu, Cora, and De~Freitas]{brochu2010tutorial}
Eric Brochu, Vlad~M Cora, and Nando De~Freitas.
\newblock A tutorial on {Bayesian optimization} of expensive cost functions,
  with application to active user modeling and hierarchical reinforcement
  learning.
\newblock \emph{arXiv preprint arXiv:1012.2599}, 2010.

\bibitem[Calandra et~al.(2016)Calandra, Seyfarth, Peters, and
  Deisenroth]{calandra2016bayesian}
Roberto Calandra, Andr{\'e} Seyfarth, Jan Peters, and Marc~Peter Deisenroth.
\newblock Bayesian optimization for learning gaits under uncertainty.
\newblock \emph{Annals of Mathematics and Artificial Intelligence}, 76\penalty0
  (1):\penalty0 5--23, 2016.

\bibitem[Derksen and Keselman(1992)]{derksen1992backward}
Shelley Derksen and Harvey~J Keselman.
\newblock Backward, forward and stepwise automated subset selection algorithms:
  Frequency of obtaining authentic and noise variables.
\newblock \emph{British Journal of Mathematical and Statistical Psychology},
  45\penalty0 (2):\penalty0 265--282, 1992.

\bibitem[Djolonga et~al.(2013)Djolonga, Krause, and Cevher]{djolonga2013high}
Josip Djolonga, Andreas Krause, and Volkan Cevher.
\newblock High-dimensional {Gaussian} process bandits.
\newblock In \emph{Advances in Neural Information Processing Systems}, pages
  1025--1033, 2013.

\bibitem[Eriksson and Jankowiak(2021)]{eriksson2021high}
David Eriksson and Martin Jankowiak.
\newblock {High-Dimensional Bayesian Optimization with Sparse Axis-Aligned
  Subspaces}.
\newblock \emph{arXiv preprint arXiv:2103.00349}, 2021.

\bibitem[Frazier(2018)]{frazier2018tutorial}
Peter~I Frazier.
\newblock A tutorial on {Bayesian} optimization.
\newblock \emph{arXiv preprint arXiv:1807.02811}, 2018.

\bibitem[Gonzalez et~al.(2015)Gonzalez, Longworth, James, and
  Lawrence]{gonzalez2015bayesian}
Javier Gonzalez, Joseph Longworth, David~C James, and Neil~D Lawrence.
\newblock Bayesian optimization for synthetic gene design.
\newblock \emph{arXiv preprint arXiv:1505.01627}, 2015.

\bibitem[Griffiths and Hern{\'a}ndez-Lobato(2017)]{griffiths2017constrained}
Ryan-Rhys Griffiths and Jos{\'e}~Miguel Hern{\'a}ndez-Lobato.
\newblock Constrained {Bayesian} optimization for automatic chemical design.
\newblock \emph{arXiv preprint arXiv:1709.05501}, 2017.

\bibitem[Hansen(2016)]{hansen2016cma}
Nikolaus Hansen.
\newblock The {CMA} evolution strategy: A tutorial.
\newblock \emph{arXiv preprint arXiv:1604.00772}, 2016.

\bibitem[Hutter et~al.(2010)Hutter, Hoos, and
  Leyton-Brown]{hutter2010automated}
Frank Hutter, Holger~H Hoos, and Kevin Leyton-Brown.
\newblock Automated configuration of mixed integer programming solvers.
\newblock In \emph{International Conference on Integration of Artificial
  Intelligence (AI) and Operations Research (OR) Techniques in Constraint
  Programming}, pages 186--202. Springer, 2010.

\bibitem[Hutter et~al.(2014)Hutter, Hoos, and
  Leyton-Brown]{hutter2014efficient}
Frank Hutter, Holger Hoos, and Kevin Leyton-Brown.
\newblock An efficient approach for assessing hyperparameter importance.
\newblock In \emph{International Conference on Machine Learning}, pages
  754--762. PMLR, 2014.

\bibitem[Jones(2008)]{jones2008large}
Donald~R Jones.
\newblock Large-scale multi-disciplinary mass optimization in the auto
  industry.
\newblock In \emph{MOPTA 2008 Conference (20 August 2008)}, 2008.

\bibitem[Kandasamy et~al.(2015)Kandasamy, Schneider, and
  P{\'o}czos]{kandasamy2015high}
Kirthevasan Kandasamy, Jeff Schneider, and Barnab{\'a}s P{\'o}czos.
\newblock High dimensional {Bayesian} optimisation and bandits via additive
  models.
\newblock In \emph{International Conference on Machine Learning}, pages
  295--304, 2015.

\bibitem[Kandasamy et~al.(2020)Kandasamy, Vysyaraju, Neiswanger, Paria,
  Collins, Schneider, Poczos, and Xing]{kandasamy2020tuning}
Kirthevasan Kandasamy, Karun~Raju Vysyaraju, Willie Neiswanger, Biswajit Paria,
  Christopher~R Collins, Jeff Schneider, Barnabas Poczos, and Eric~P Xing.
\newblock Tuning hyperparameters without grad students: Scalable and robust
  bayesian optimisation with dragonfly.
\newblock \emph{Journal of Machine Learning Research}, 21\penalty0
  (81):\penalty0 1--27, 2020.

\bibitem[Klein et~al.(2017)Klein, Falkner, Bartels, Hennig, and
  Hutter]{klein2017fast}
Aaron Klein, Stefan Falkner, Simon Bartels, Philipp Hennig, and Frank Hutter.
\newblock Fast {Bayesian} optimization of machine learning hyperparameters on
  large datasets.
\newblock In \emph{Artificial Intelligence and Statistics}, pages 528--536.
  PMLR, 2017.

\bibitem[Letham et~al.(2020)Letham, Calandra, Rai, and Bakshy]{letham2020re}
Ben Letham, Roberto Calandra, Akshara Rai, and Eytan Bakshy.
\newblock Re-examining linear embeddings for high-dimensional {Bayesian}
  optimization.
\newblock \emph{Advances in Neural Information Processing Systems}, 33, 2020.

\bibitem[Li et~al.(2016)Li, Kandasamy, P{\'o}czos, and Schneider]{li2016high}
Chun-Liang Li, Kirthevasan Kandasamy, Barnab{\'a}s P{\'o}czos, and Jeff
  Schneider.
\newblock High dimensional {Bayesian} optimization via restricted projection
  pursuit models.
\newblock In \emph{Artificial Intelligence and Statistics}, pages 884--892,
  2016.

\bibitem[Loizou and Richt{\'a}rik(2017)]{loizou2017momentum}
Nicolas Loizou and Peter Richt{\'a}rik.
\newblock Momentum and stochastic momentum for stochastic gradient, {Newton},
  proximal point and subspace descent methods.
\newblock \emph{arXiv preprint arXiv:1712.09677}, 2017.

\bibitem[Marco et~al.(2017)Marco, Berkenkamp, Hennig, Schoellig, Krause,
  Schaal, and Trimpe]{marco2017virtual}
Alonso Marco, Felix Berkenkamp, Philipp Hennig, Angela~P Schoellig, Andreas
  Krause, Stefan Schaal, and Sebastian Trimpe.
\newblock Virtual vs. real: Trading off simulations and physical experiments in
  reinforcement learning with {Bayesian} optimization.
\newblock In \emph{2017 IEEE International Conference on Robotics and
  Automation (ICRA)}, pages 1557--1563. IEEE, 2017.

\bibitem[Metzen(2016)]{metzen2016minimum}
Jan~Hendrik Metzen.
\newblock Minimum regret search for single- and multi-task optimization.
\newblock \emph{arXiv preprint arXiv:1602.01064}, 2016.

\bibitem[Mo{\v{c}}kus(1975)]{movckus1975bayesian}
Jonas Mo{\v{c}}kus.
\newblock On {Bayesian} methods for seeking the extremum.
\newblock In \emph{Optimization Techniques IFIP Technical Conference}, pages
  400--404. Springer, 1975.

\bibitem[Moriconi et~al.(2019)Moriconi, Deisenroth, and
  Kumar]{moriconi2019high}
Riccardo Moriconi, Marc~P Deisenroth, and KS~Kumar.
\newblock High-dimensional {Bayesian} optimization using low-dimensional
  feature spaces.
\newblock \emph{arXiv preprint arXiv:1902.10675}, 2019.

\bibitem[Nayebi et~al.(2019)Nayebi, Munteanu, and
  Poloczek]{nayebi2019framework}
Amin Nayebi, Alexander Munteanu, and Matthias Poloczek.
\newblock A framework for {Bayesian} optimization in embedded subspaces.
\newblock In \emph{International Conference on Machine Learning}, pages
  4752--4761. PMLR, 2019.

\bibitem[Negoescu et~al.(2011)Negoescu, Frazier, and
  Powell]{negoescu2011knowledge}
Diana~M Negoescu, Peter~I Frazier, and Warren~B Powell.
\newblock The knowledge-gradient algorithm for sequencing experiments in drug
  discovery.
\newblock \emph{INFORMS Journal on Computing}, 23\penalty0 (3):\penalty0
  346--363, 2011.

\bibitem[Nickson et~al.(2014)Nickson, Osborne, Reece, and
  Roberts]{nickson2014automated}
Thomas Nickson, Michael~A Osborne, Steven Reece, and Stephen~J Roberts.
\newblock Automated machine learning on big data using stochastic algorithm
  tuning.
\newblock \emph{arXiv preprint arXiv:1407.7969}, 2014.

\bibitem[Paananen et~al.(2019)Paananen, Piironen, Andersen, and
  Vehtari]{paananen2019variable}
Topi Paananen, Juho Piironen, Michael~Riis Andersen, and Aki Vehtari.
\newblock Variable selection for gaussian processes via sensitivity analysis of
  the posterior predictive distribution.
\newblock In \emph{The 22nd International Conference on Artificial Intelligence
  and Statistics}, pages 1743--1752, 2019.

\bibitem[Rolland et~al.(2018)Rolland, Scarlett, Bogunovic, and
  Cevher]{rolland2018high}
Paul Rolland, Jonathan Scarlett, Ilija Bogunovic, and Volkan Cevher.
\newblock High-dimensional {Bayesian} optimization via additive models with
  overlapping groups.
\newblock \emph{arXiv preprint arXiv:1802.07028}, 2018.

\bibitem[Snoek et~al.(2012)Snoek, Larochelle, and Adams]{snoek2012practical}
Jasper Snoek, Hugo Larochelle, and Ryan~P Adams.
\newblock Practical {Bayesian} optimization of machine learning algorithms.
\newblock In \emph{Advances in Neural Information Processing Systems}, pages
  2951--2959, 2012.

\bibitem[Spagnol et~al.(2019)Spagnol, Le~Riche, and
  Da~Veiga]{spagnol2019bayesian}
Adrien Spagnol, Rodolphe Le~Riche, and S{\'e}bastien Da~Veiga.
\newblock Bayesian optimization in effective dimensions via kernel-based
  sensitivity indices.
\newblock In \emph{International Conference on Applications of Statistics and
  Probability in Civil Engineering}, 2019.

\bibitem[Srinivas et~al.(2009)Srinivas, Krause, Kakade, and
  Seeger]{srinivas2009gaussian}
Niranjan Srinivas, Andreas Krause, Sham~M Kakade, and Matthias Seeger.
\newblock Gaussian process optimization in the bandit setting: No regret and
  experimental design.
\newblock \emph{arXiv preprint arXiv:0912.3995}, 2009.

\bibitem[Stewart(1980)]{stewart1980efficient}
Gilbert~W Stewart.
\newblock The efficient generation of random orthogonal matrices with an
  application to condition estimators.
\newblock \emph{SIAM Journal on Numerical Analysis}, 17\penalty0 (3):\penalty0
  403--409, 1980.

\bibitem[Ulmasov et~al.(2016)Ulmasov, Baroukh, Chachuat, Deisenroth, and
  Misener]{ulmasov2016bayesian}
Doniyor Ulmasov, Caroline Baroukh, Benoit Chachuat, Marc~Peter Deisenroth, and
  Ruth Misener.
\newblock Bayesian optimization with dimension scheduling: Application to
  biological systems.
\newblock In \emph{Computer Aided Chemical Engineering}, volume~38, pages
  1051--1056. Elsevier, 2016.

\bibitem[Wang et~al.(2017)Wang, Li, Jegelka, and Kohli]{wang2017batched}
Zi~Wang, Chengtao Li, Stefanie Jegelka, and Pushmeet Kohli.
\newblock Batched high-dimensional bayesian optimization via structural kernel
  learning.
\newblock In \emph{International Conference on Machine Learning}, pages
  3656--3664. PMLR, 2017.

\bibitem[Wang et~al.(2016)Wang, Hutter, Zoghi, Matheson, and
  de~Feitas]{wang2016bayesian}
Ziyu Wang, Frank Hutter, Masrour Zoghi, David Matheson, and Nando de~Feitas.
\newblock Bayesian optimization in a billion dimensions via random embeddings.
\newblock \emph{Journal of Artificial Intelligence Research}, 55:\penalty0
  361--387, 2016.

\bibitem[Williams and Rasmussen(2006)]{williams2006gaussian}
Christopher~KI Williams and Carl~Edward Rasmussen.
\newblock \emph{Gaussian processes for machine learning}, volume~2.
\newblock MIT press Cambridge, MA, 2006.

\bibitem[Wilson et~al.(2014)Wilson, Fern, and Tadepalli]{wilson2014using}
Aaron Wilson, Alan Fern, and Prasad Tadepalli.
\newblock Using trajectory data to improve {Bayesian} optimization for
  reinforcement learning.
\newblock \emph{The Journal of Machine Learning Research}, 15\penalty0
  (1):\penalty0 253--282, 2014.

\bibitem[Yao et~al.(2018)Yao, Wang, Chen, Dai, Yi-Qi, Yu-Feng, Wei-Wei, Qiang,
  and Yang]{yao2018taking}
Quanming Yao, Mengshuo Wang, Yuqiang Chen, Wenyuan Dai, Hu~Yi-Qi, Li~Yu-Feng,
  Tu~Wei-Wei, Yang Qiang, and Yu~Yang.
\newblock Taking human out of learning applications: A survey on automated
  machine learning.
\newblock \emph{arXiv preprint arXiv:1810.13306}, 2018.

\end{thebibliography}

%References follow the acknowledgments. Use unnumbered first-level heading for
%the references. Any choice of citation style is acceptable as long as you are
%consistent. It is permissible to reduce the font size to \verb+small+ (9 point)
%when listing the references.
%Note that the Reference section does not count towards the page limit.
\medskip

%{
%\small

%[1] Alexander, J.A.\ \& Mozer, M.C.\ (1995) Template-based algorithms for
%connectionist rule extraction. In G.\ Tesauro, D.S.\ Touretzky and T.K.\ Leen
%(eds.), {\it Advances in Neural Information Processing Systems 7},
%pp.\ 609--616. Cambridge, MA: MIT Press.

%[2] Bower, J.M.\ \& Beeman, D.\ (1995) {\it The Book of GENESIS: Exploring
 % Realistic Neural Models with the GEneral NEural SImulation System.}  New York:
%TELOS/Springer--Verlag.

%[3] Hasselmo, M.E., Schnell, E.\ \& Barkai, E.\ (1995) Dynamics of learning and
%recall at excitatory recurrent synapses and cholinergic modulation in rat
%hippocampal region CA3. {\it Journal of Neuroscience} {\bf 15}(7):5249-5262.
%}

%%%%%%%%%%%%%%%%%%%%%%%%%%%%%%%%%%%%%%%%%%%%%%%%%%%%%%%%%%%%
\section*{Checklist}

%%% BEGIN INSTRUCTIONS %%%
\begin{comment}

The checklist follows the references.  Please
read the checklist guidelines carefully for information on how to answer these
questions.  For each question, change the default \answerTODO{} to \answerYes{},
\answerNo{}, or \answerNA{}.  You are strongly encouraged to include a {\bf
justification to your answer}, either by referencing the appropriate section of
your paper or providing a brief inline description.  For example:
\begin{itemize}
  \item Did you include the license to the code and datasets? \answerYes{See Section~\ref{gen_inst}.}
  \item Did you include the license to the code and datasets? \answerNo{The code and the data are proprietary.}
  \item Did you include the license to the code and datasets? \answerNA{}
\end{itemize}
Please do not modify the questions and only use the provided macros for your
answers.  Note that the Checklist section does not count towards the page
limit.  In your paper, please delete this instructions block and only keep the
Checklist section heading above along with the questions/answers below.
\end{comment}
%%% END INSTRUCTIONS %%%

\begin{enumerate}

\item For all authors...
\begin{enumerate}
  \item Do the main claims made in the abstract and introduction accurately reflect the paper's contributions and scope?
    \answerYes{} The last paragraph of the introduction section describes the paper's contributions.
  \item Did you describe the limitations of your work?
    \answerYes{} The last paragraph of the conclusion section describes some limitations of our work.
  \item Did you discuss any potential negative societal impacts of your work?
    \answerNA{} There will be no potential negative societal impacts of our work. 
  \item Have you read the ethics review guidelines and ensured that your paper conforms to them?
    \answerYes{} I read it carefully. 
\end{enumerate}

\item If you are including theoretical results...
\begin{enumerate}
  \item Did you state the full set of assumptions of all theoretical results?
    \answerYes{} It is in section 5. 
	\item Did you include complete proofs of all theoretical results?
     \answerYes{} It is in section B and C. 
\end{enumerate}

\item If you ran experiments...
\begin{enumerate}
  \item Did you include the code, data, and instructions needed to reproduce the main experimental results (either in the supplemental material or as a URL)?
    \answerYes{} Codes will be attached in the supplementary material. 
  \item Did you specify all the training details (e.g., data splits, hyperparameters, how they were chosen)?
   \answerYes{} They are described in section D. 
	\item Did you report error bars (e.g., with respect to the random seed after running experiments multiple times)?
    \answerYes{} For each case we repeat $20$ independent runs, and we plot mean and standard deviation in our figures. 
	\item Did you include the total amount of compute and the type of resources used (e.g., type of GPUs, internal cluster, or cloud provider)?
    \answerYes{} They are described in section D. 
\end{enumerate}

\item If you are using existing assets (e.g., code, data, models) or curating/releasing new assets...
\begin{enumerate}
  \item If your work uses existing assets, did you cite the creators?
    \answerYes{}
  \item Did you mention the license of the assets?
    \answerNA{}
  \item Did you include any new assets either in the supplemental material or as a URL?
    \answerYes{} They are described in section D. 
  \item Did you discuss whether and how consent was obtained from people whose data you're using/curating?
     \answerNA{}
  \item Did you discuss whether the data you are using/curating contains personally identifiable information or offensive content?
    \answerNA{}
\end{enumerate}

\item If you used crowdsourcing or conducted research with human subjects...
\begin{enumerate}
  \item Did you include the full text of instructions given to participants and screenshots, if applicable?
    \answerNA{}
  \item Did you describe any potential participant risks, with links to Institutional Review Board (IRB) approvals, if applicable?
    \answerNA{}
  \item Did you include the estimated hourly wage paid to participants and the total amount spent on participant compensation?
    \answerNA{}
\end{enumerate}

\end{enumerate}

%%%%%%%%%%%%%%%%%%%%%%%%%%%%%%%%%%%%%%%%%%%%%%%%%%%%%%%%%%%%

\newpage
\appendix

\section{Variable selection with momentum mechanism}

In this section, we provide pseudo-code for the algorithm of variable selection with momentum mechanism. Note that Algorithm~\ref{alg:momentum_inacc} (momentum in the inaccurate case) is similar to Algorithm~\ref{alg:VS_detail}, except the lines that are marked with red color. 

\begin{algorithm}
    \caption{Variable Selection (VS) with Momentum} 
    \begin{algorithmic}[1]
        \State \textbf{Input}: iteration index $t$, $\mathcal{D}=\{(\mathbf{x}^{i},y^{i})\}_{i=1}^{t}$, $N_{init}$, $N_{vs}$, set of important variables chosen at iteration $t-N_{vs}$, denote as $\hat{\mathbf{x}}_{ipt}$
        \State \textbf{Output}: Set of important variables chosen at iteration $t$, denote as $\mathbf{x}_{ipt}$
	    \If{$t=N_{init}+N_{vs}$ or $\hat{\mathbf{x}}_{ipt}=\mathbf{x}$} \Comment{First time to do variable selection or $\hat{\mathbf{x}}_{ipt}$ contains all variables}
	        \State \Return Algorithm~\ref{alg:VS_detail}
	   \ElsIf{$\max_{k\in \{t-N_{vs}+1, t-N_{vs}+2, \dots, t\}}y^{k}\leq\max_{k\in \{1,\dots, t-N_{vs}\}}y^{k}$}  \Comment{Inaccurate case}
	        \State \Return Algorithm~\ref{alg:momentum_inacc}
	   \Else \Comment{Accurate case}
	   %\ElsIf{$\max_{k\in \{T-N_{vs}+1, T-N_{vs}+2, \dots, T\}}y^{k}>\max_{k\in \{1,\dots, T-N_{vs}\}}y^{k}$} 
	        \State \Return Algorithm~\ref{alg:momentum_acc}
	    \EndIf
    \end{algorithmic} 
    \label{alg:VS_momentum}
\end{algorithm}

\begin{algorithm}
	\caption{Momentum in the inaccurate case} 
	\begin{algorithmic}[1]
	    \State \textbf{Input}: $\mathcal{D}=\{(\mathbf{x}^{i},y^{i})\}_{i=1}^{t}$, $N_{vs}$, set of important variables chosen at iteration $t-N_{vs}$, denote as $\hat{\mathbf{x}}_{ipt}$
	    \State \textbf{Output}: Set of important variables chosen at iteration $t$, denote as $\mathbf{x}_{ipt}$
	    \State Fit a GP to $\mathcal{D}$ and calculate important scores of variables $IS$ where $IS[i]$ is the important score of the i-th variable 
	    \State Sort variables according to their important scores, $[\mathbf{x}_{s(1)},\dots, \mathbf{x}_{s(D)}]$, from the most important to the least
	    %\State Fit GP using the most important variables: $Y=(y^{1:t})\sim \mathcal{N}(\mathbf{0},K(x^{1:t}_{s(1)},\Theta)+\sigma^{2}_{t}I)$, maximize the marginal likelihood and record the final loss $L_1$.
    \For{$n=1,\ldots D$}
	        \textcolor{red}{\If{$\mathbf{x}_{s(n)}\notin \hat{\mathbf{x}}_{ipt}$}
	            \State \textbf{break}
	        \EndIf}
	    \EndFor
	    \For{$m=n,n+1,\ldots D$} \Comment{Stepwise forward selection}
	        \State Fit a GP to $\mathcal{D}_{m}:=\{(\mathbf{x}_{s(1):s(m)}^{i},y^{i})\}_{i=1}^{t-1}$ where $\mathbf{x}_{s(1):s(m)}^{i}$ is the $i$-th input with only the first $m$ important variables, let $L_m$ to be the value of final negative marginal log likelihood
	        \If{$m-n<2$}
	            \State \textbf{continue}
	        %\State Fit GP using the first $m$ important variables: $Y=(y^{1:t})\sim \mathcal{N}(\mathbf{0},K(x^{1:t}_{s(1):s(m)},\Theta)+\sigma^{2}_{t}I)$, maximize the marginal likelihood and record the final loss $L_m$.
	        \ElsIf {$L_{m-1}-L_{m}\leq 0$ or $L_{m-1}-L_{m}<\frac{L_{m-2}-L(m-1)}{10}$}
	            \State \textbf{break}
	        \EndIf
	    \EndFor
	    \State\Return
	    $\mathbf{x}_{ipt}=\{\mathbf{x}_{s(1)},\dots, \mathbf{x}_{s(m-1)}\}$
	\end{algorithmic} 
	\label{alg:momentum_inacc}
\end{algorithm}

\begin{algorithm}
	\caption{Momentum in accurate case} 
	\begin{algorithmic}[1]
	    \State \textbf{Input}: $\mathcal{D}=\{(\mathbf{x}^{i},y^{i})\}_{i=1}^{t}$, $N_{vs}$, set of important variables chosen at iteration $t-N_{vs}$, denoted as $\hat{\mathbf{x}}_{ipt}$, with cardinality $w=\abs{\hat{\mathbf{x}}_{ipt}}$
	    \State \textbf{Output}: Set of important variables chosen at iteration $t$, denote as $\mathbf{x}_{ipt}$
	    %\State Denote $I_{ipt}^{T-N_{vs}}$ be indices of important variables selected at $T-N_{vs}$.
	    \State Fit a GP to $\mathcal{D}$ and calculate important scores of variables $IS$ where $IS[i]$ is the important score of the i-th variable 
	    %\State Fit GP using all variables: $Y=(y^{1:T})\sim \mathcal{N}(\mathbf{0},K(x^{1:T},\Theta)+\sigma^{2}_{T}I)$, maximize the marginal likelihood and record the final loss $L_0$.
	    %\State Calculate important scores of variables, $IS$, $IS[i]$ is the important score of the i-th variable. 
	    \State Sort variables according to $IS$, $[\mathbf{x}_{s(1)},\dots, \mathbf{x}_{s(D)}]$, from the most important to the least
	    %\State Fit GP using the most important variables: $Y=(y^{1:T})\sim \mathcal{N}(\mathbf{0},K(x^{1:T}_{s(1)},\Theta)+\sigma^{2}_{T}I)$, maximize the marginal likelihood and record the final loss $L_1$.
	    \State Fit a GP by using variables in $\hat{\mathbf{x}}_{ipt}$, i.e. fit a GP to $\{(\hat{\mathbf{x}}_{ipt}^{i},y^{i})\}_{i=1}^{t}$, and calculate important scores of these variables $\widehat{IS}$. Let $\hat{L}_{w}$ be the value of final negative marginal log likelihood
        %\State Fit GP using variables in $I_{ipt}^{T-N_{vs}}$: $Y=(y^{1:t})\sim \mathcal{N}(\mathbf{0},K(x^{1:t}_{I_{ipt}^{T-N_{vs}}},\Theta)+\sigma^{2}_{t}I)$, maximize the marginal likelihood and record the final loss $L_{ipt}^{T-N_{vs}}$.
        %\State Calculate important scores of variables in $I_{ipt}^{T-N_{vs}}$, $IS_{ipt}^{T-N_{vs}}$, $IS_{ipt}^{T-N_{vs}}[i]$ is the important score of the i-th variable in $I_{ipt}^{T-N_{vs}}$
        \State Sort variables in $\hat{\mathbf{x}}_{ipt}$ according to $\widehat{IS}$, $[\mathbf{x}_{s'(1)},\dots, \mathbf{x}_{s'(w)}]$, from the most important to the least.
        \For{$m=w-1,w-2,\ldots 0$} \Comment{Recursive feature elimination}
            \If{$m=0$}
                \State Set $\mathbf{x}_{ipt}=\{\mathbf{x}_{s'(1)}\}$
                \State \textbf{break}
            \EndIf
            \State Fit a GP by only using the first $m$ important variables according to $\widehat{IS}$. Let $\hat{L}_m$ to be the value of final negative marginal log likelihood
            \If{$\hat{L}_m>\hat{L}_{m+1}$}
                \State Set $\mathbf{x}_{ipt}=\{\mathbf{x}_{s'(1)},\dots, \mathbf{x}_{s'(m+1)}\}$
                \State Set $L_{0}=\hat{L}_{m+1}$
                \State \textbf{break}
            \EndIf
        \EndFor
        \For{$m=1,2,\ldots D$} \Comment{Stepwise forward selection}
            \If{$\mathbf{x}_{s(m)}\in \mathbf{x}_{ipt}$}
                \State Set $L_{m}=L_{m-1}$, $L_{m-1}=L_{m-2}$
                \State \textbf{continue}
            \EndIf
            \State Fit a GP by using variables in $\mathbf{x}_{ipt}\cup \{\mathbf{x}_{s(m)}\}$. Let $L_{m}$ to be the value of final negative marginal log likelihood
            \If{$m<2$}
                \State $\mathbf{x}_{ipt}=\mathbf{x}_{ipt}\cup \{\mathbf{x}_{s(m)}\}$
	            \State \textbf{continue}
	        %\State Fit GP using the first $m$ important variables: $Y=(y^{1:t})\sim \mathcal{N}(\mathbf{0},K(x^{1:t}_{s(1):s(m)},\Theta)+\sigma^{2}_{t}I)$, maximize the marginal likelihood and record the final loss $L_m$.
	        \ElsIf {$L_{m-1}-L_{m}\leq 0$ or $L_{m-1}-L_{m}<\frac{L_{m-2}-L(m-1)}{10}$}
	            \State \textbf{break}
	        \EndIf
	        \State $\mathbf{x}_{ipt}=\mathbf{x}_{ipt}\cup \{\mathbf{x}_{s(m)}\}$
        \EndFor
	    \State \Return $\mathbf{x}_{ipt}$
	\end{algorithmic} 
	\label{alg:momentum_acc}
\end{algorithm}

\section{Proof of Proposition \ref{prop:complexity}}

\begin{proof}[Proof of Proposition \ref{prop:complexity}] Given query-output pairs  $\mathcal{D}=\{(\mathbf{x}^{i},y^{i})\}_{i=1}^{n}$, the marginal log likelihood (MLL) that need to be maximized at the step of fitting a GP has the following explicit form:
\begin{align*}
    \log p(\Theta=\{\rho^{2}_{1:D},\alpha_{0}^{2}\},\sigma_{0}\mid \mathcal{D}) 
    %&= -\frac{1}{2}\mathbf{y}\left(K(\mathbf{x}^{1:n},\Theta)+\sigma^{2}_{0}\mathbf{I}\right)^{-1}\mathbf{y}^{\top} - \frac{1}{2}\log\abs{K(\mathbf{x}^{1:n},\Theta)+\sigma^{2}_{0}\mathbf{I}}-\frac{n\log2\pi}{2} \\
     = -\frac{1}{2}\mathbf{y}M^{-1}\mathbf{y}^{\top} - \frac{1}{2}\log\abs{M}-\frac{n\log2\pi}{2}
\end{align*}
where $\mathbf{y}=[y^{1},\ldots y^{n}]$ is an $n$-dimensional vector and $M=\left(K(\mathbf{x}^{1:n},\Theta)+\sigma^{2}_{0}\mathbf{I}\right)$. When the quasi-Newton method is used for maximizing MLL, the gradient should be calculated for each iteration:
\begin{align*}
    \bigtriangledown_{\Theta,\sigma_{0}}\log  p(\Theta,\sigma_{0}\mid \mathcal{D}) = -\frac{1}{2}\mathbf{y}M^{-1}\left(\bigtriangledown_{\Theta,\sigma_{0}}M\right)M^{-1}\mathbf{y}^{\top}-\frac{1}{2}\textbf{tr}\left(M^{-1}\left(\bigtriangledown_{\Theta,\sigma_{0}}M\right)\right)
\end{align*}
When only variables in $\mathbf{x}_{ipt}$ are used, we define the distance between two queries $\mathbf{x}^{i}$ and $\mathbf{x}^{j}$ as:
\begin{align*}
    d(\mathbf{x}^{i},\mathbf{x}^{j})=\sqrt{\sum_{m: m\in\mathbf{x}_{ipt}}\rho_{m}^{2}(\mathbf{x}_{m}^{i}-\mathbf{x}_{m}^{j})^2}
\end{align*}
and all the other inverse squared length scales corresponding to unimportant variables are fixed to $0$. Commonly chosen kernel functions are actually functions of the distance defined above, for example the squared exponential (SE) kernel is as the following:
\begin{align*}
    k_{SE}(\mathbf{x}^{i},\mathbf{x}^{j},\Theta)=\alpha_{0}^{2}\exp\left(-\frac{1}{2}d^{2}(\mathbf{x}^{i},\mathbf{x}^{j})\right)
\end{align*}
and the Matern-5/2 kernel is as the following:
\begin{align*}
    k_{Mt}(\mathbf{x}^{i},\mathbf{x}^{j},\Theta)=\alpha_{0}^{2}\left(1+\sqrt{5}d(\mathbf{x}^{i},\mathbf{x}^{j})+\frac{5}{3}d^{2}(\mathbf{x}^{i},\mathbf{x}^{j})\right)\exp\left(-\sqrt{5}d(\mathbf{x}^{i},\mathbf{x}^{j})\right)
\end{align*}
Since the cardinality of $\mathbf{x}_{ipt}$ is $p$, the cardinality of parameters in the kernel function that are not fixed to $0$ is $p+1$, hence the complexity of calculating the gradient of the distance is $\mathcal{O}(p)$, therefore whatever using SE kernel or Matern-5/2 kernel, the complexity of calculating $\bigtriangledown_{\Theta}k(\mathbf{x}^{i},\mathbf{x}^{j},\Theta)$ is $\mathcal{O}(p)$. 

Since $M$ is a $n\times n$ matrix and each entry $M_{ij}$ equals to $k(\mathbf{x}^{i},\mathbf{x}^{j},\Theta)+\sigma_{0}^{2}\mathbbm{1}(i=j)$, the complexity of calculating $\bigtriangledown_{\Theta,\sigma_{0}}M$ is $\mathcal{O}(pn^{2})$. the complexity of calculating the inverse matrix $M^{-1}$ is $\mathcal{O}(n^{3})$ in general, and the following matrix multiplication and trace calculation need $\mathcal{O}(pn^{2})$, therefore the complexity of calculating the gradient of MLL is $\mathcal{O}(pn^2+n^3)$. Once the gradient is obtained, each quasi-Newton step needs additional $\mathcal{O}(p^2)$, therefore the complexity of one step of quasi-Newton method when fitting a GP is $\mathcal{O}(p^2+pn^2+n^3)$. 

As described in section 2, the acquisition function is a function that depends on the posterior mean $\mu$ and the posterior standard deviation $\sigma$, hence the gradients of $\mu$ and $\sigma$ should be calculated when the gradient of the acquisition function is needed. 

When only variables in $\mathbf{x}_{ipt}$ are used, the gradient of $\mu$ with respect to $\mathbf{x}_{ipt}$ has the following form:
\begin{align*}
    \bigtriangledown_{\mathbf{x}_{ipt}}\mu(\mathbf{x}_{ipt}\mid \mathcal{D})=\left(\bigtriangledown_{\mathbf{x}_{ipt}}K(\mathbf{x}_{ipt},\mathbf{x}^{1:n}_{ipt})\right)\left(K(\mathbf{x}^{1:n}_{ipt},\Theta)+\sigma^{2}_{0}\mathbf{I}\right)^{-1}\mathbf{y}^{\top}
\end{align*}

Here $\left(K(\mathbf{x}^{1:n}_{ipt},\Theta)+\sigma^{2}_{0}\mathbf{I}\right)^{-1}\mathbf{y}^{\top}$ is fixed so that its value can be calculated in advance and stored as a $n$-dimensional vector. $K(\mathbf{x}_{ipt},\mathbf{x}^{1:n}_{ipt})$ is a $n$-dimensional vector of which each element is a kernel value between $\mathbf{x}_{ipt}$ and $\mathbf{x}_{ipt}^{i}$, hence the complexity of calculating the gradient of each element in $K(\mathbf{x}_{ipt},\mathbf{x}^{1:n}_{ipt})$ is $\mathcal{O}(p)$. Therefore, the complexity is $\mathcal{O}(pn)$ to calculate $\bigtriangledown_{\mathbf{x}_{ipt}}K(\mathbf{x}_{ipt},\mathbf{x}^{1:n}_{ipt})$ and $\mathcal{O}(pn)$ for additional matrix manipulation, hence the total complexity for calculating $\bigtriangledown_{\mathbf{x}_{ipt}}\mu(\mathbf{x}_{ipt}\mid \mathcal{D})$ is $\mathcal{O}(pn)$. 

The gradient of $\sigma$ has the following form:
\begin{align*}
    \bigtriangledown_{\mathbf{x}_{ipt}}\sigma(\mathbf{x}_{ipt}\mid \mathcal{D}) &=\bigtriangledown_{\mathbf{x}_{ipt}}\sqrt{k(\mathbf{x}_{ipt},\mathbf{x}_{ipt},\Theta)-K(\mathbf{x}_{ipt},\mathbf{x}_{ipt}^{1:n})[K(\mathbf{x}_{ipt}^{1:n},\Theta)+\sigma^{2}_{0}\mathbf{I}]^{-1}K(\mathbf{x}_{ipt},\mathbf{x}_{ipt}^{1:n})^{\top}} \\
    &=-\frac{\left(\bigtriangledown_{\mathbf{x}_{ipt}}K(\mathbf{x}_{ipt},\mathbf{x}^{1:n}_{ipt})\right)\left(K(\mathbf{x}^{1:n}_{ipt},\Theta)+\sigma^{2}_{0}\mathbf{I}\right)^{-1}}{\sqrt{k(\mathbf{x}_{ipt},\mathbf{x}_{ipt},\Theta)-K(\mathbf{x}_{ipt},\mathbf{x}_{ipt}^{1:n})[K(\mathbf{x}_{ipt}^{1:n},\Theta)+\sigma^{2}_{0}\mathbf{I}]^{-1}K(\mathbf{x}_{ipt},\mathbf{x}_{ipt}^{1:n})^{\top}}}
\end{align*}
Once $\bigtriangledown_{\mathbf{x}_{ipt}}K(\mathbf{x}_{ipt},\mathbf{x}^{1:n}_{ipt})$ is calculated, $\mathcal{O}(pn+n^2)$ is needed for additional matrix manipulation, hence the total complexity for calculating $ \bigtriangledown_{\mathbf{x}_{ipt}}\sigma(\mathbf{x}_{ipt}\mid \mathcal{D})$ is $\mathcal{O}(pn+n^2)$. 

For commonly used acquisition functions such as upper confidence bound (UCB)~\citep{auer2002using}:
\begin{align*}
    UCB(\mathbf{x}_{ipt}\mid \mathcal{D}) = \mu(\mathbf{x}_{ipt}\mid \mathcal{D}) + \sqrt{\beta_{n}}\sigma(\mathbf{x}_{ipt}\mid \mathcal{D})
\end{align*}
and expected improvement (EI)~\citep{movckus1975bayesian}:
\begin{align*}
    EI(\mathbf{x}_{ipt}\mid \mathcal{D})=\left(\mu(\mathbf{x}_{ipt}\mid \mathcal{D})-y^{*}_{n}\right)\Phi\left(\frac{\mu(\mathbf{x}_{ipt}\mid \mathcal{D})-y^{*}_{n}}{\sigma(\mathbf{x}_{ipt}\mid \mathcal{D})}\right) + \sigma(\mathbf{x}_{ipt}\mid \mathcal{D})\varphi\left(\frac{\mu(\mathbf{x}_{ipt}\mid \mathcal{D})-y^{*}_{n}}{\sigma(\mathbf{x}_{ipt}\mid \mathcal{D})}\right)
\end{align*}
where $y_{n}^{*}=\max_{i\leq n} y^{i}$, $\Phi(\cdot)$ is the cumulative distribution function of the standard normal distribution, and $\varphi(\cdot)$ is the probability density function, once the gradients of $\mu$ and $\sigma$ are derived, only additional $\mathcal{O}(p)$ is needed for vector calculation, hence the total complexity of calculating the gradient of the acquisition function is $\mathcal{O}(pn+n^2)$. Again, once the gradient is obtained, each quasi-Newton step needs additional $\mathcal{O}(p^2)$, therefore the complexity of one step of quasi-Newton method for maximising the acquisition function is $\mathcal{O}(p^2+pn+n^2)$. 

\end{proof}

\begin{algorithm}
	\caption{VS-GP-UCB} 
	\begin{algorithmic}[1]
	    \State \textbf{Input}: domain $\mathcal{X}$, GP prior with mean $\mu_0=0$ and the kernel function $k$, maximal iteration $N$
	    \State \textbf{Output}: Approximate maximizer $\mathbf{x}^{max}$
	    \State Let $\mathcal{D}=\emptyset$
	    \State Fix the last $D-d$ variables, $\mathbf{x}_{[d+1:D]}^{0}$
		\For {$t=1,2,\ldots N$}
		    %\State Uniformly sample the last $D-d$ variables to obtain $\mathbf{x}^{t}_{[d+1:D]}$
		    \State Choose $\mathbf{x}^{t}_{[1:d]} = \argmax_{\mathbf{x}_{[1:d]}}\mu(\{\mathbf{x}_{[1:d]},\mathbf{x}^{0}_{[d+1:D]}\}\mid\mathcal{D}) + \sqrt{\beta_t}\sigma(\{\mathbf{x}_{[1:d]},\mathbf{x}^{0}_{[d+1:D]}\}\mid\mathcal{D})$
		    \State Sample $y^{t}=f\left(\mathbf{x}^{t}=\{\mathbf{x}_{[1:d]}^{t},\mathbf{x}_{[d+1:D]}^{0}\}\right) + \epsilon_t $, where $\epsilon_t\sim\mathcal{N}(0,\sigma_{0}^2)$ is a noise
		    \State Update $\mathcal{D}=\mathcal{D}\cup \{(\mathbf{x}^{t},y^{t})\}$
		    %\State Perform Bayesian update to obtain $\mu_t$ and $\sigma_t$.
			%\For {$actor=1,2,\ldots,N$}
				%\State Run policy $\pi_{\theta_{old}}$ in environment for $T$ time steps
				%\State Compute advantage estimates $\hat{A}_{1},\ldots,\hat{A}_{T}$
			%\EndFor
			%\State Optimize surrogate $L$ wrt. $\theta$, with $K$ epochs and minibatch size $M\leq NT$
			%\State $\theta_{old}\leftarrow\theta$
		\EndFor
		\State\Return $\mathbf{x}^{max}$ that is equal to $\mathbf{x}^{i}$ with maximal $y^{i}$
	\end{algorithmic} 
	\label{alg:VS_GP_UCB}
\end{algorithm}

\section{Proof of Theorem \ref{thm:regret}}

We show the details of VS-GP-UCB and provide the proof of Theorem \ref{thm:regret} below.

\begin{proof}[Proof of Theorem \ref{thm:regret}] The second inequality of Assumption \ref{assump:derivative} is equivalent to the following inequality: 
\begin{align*}
     P(\sup_{\mathbf{x}\in \mathcal{X}}\abs{\frac{\partial f}{\partial \mathbf{x}_j}}> \alpha L) \leq a\exp\left(-\left(\frac{L}{b}\right)^2\right), j=d+1,\dots , D
\end{align*}

Therefore, according to Assumption \ref{assump:derivative} and the union bound, we have that $w.p. \geq 1-Da\exp\left(-\left(\frac{L}{b}\right)^2\right)$:
\begin{align*}
    \forall \mathbf{x}, \mathbf{x}'\in\mathcal{X}, \abs{f(\mathbf{x})-f(\mathbf{x}')}\leq L\norm{\mathbf{x}_{[1:d]}-\mathbf{x}'_{[1:d]}}_1 + \alpha L\norm{\mathbf{x}_{[d+1:D]}-\mathbf{x}'_{[d+1:D]}}_1
\end{align*}

Let $\frac{\delta}{2}=Da\exp\left(-\left(\frac{L}{b}\right)^2\right)$, meaning $L=b\sqrt{\log\left(\frac{2Da}{\delta}\right)}$, we have that $w.p. \geq 1-\frac{\delta}{2}$, 
\begin{align*}
    \forall \mathbf{x}, \mathbf{x}'\in\mathcal{X}, \abs{f(\mathbf{x})-f(\mathbf{x}')}\leq b\sqrt{\log\left(\frac{2Da}{\delta}\right)}\left(\norm{\mathbf{x}_{[1:d]}-\mathbf{x}'_{[1:d]}}_1 + \alpha \norm{\mathbf{x}_{[d+1:D]}-\mathbf{x}'_{[d+1:D]}}_1\right)
\end{align*}
%which is called smoothness inequality in the following proof. 

\begin{algorithm}
	\caption{GP-UCB (Algorithm 1 in \citet{srinivas2009gaussian})} 
	\begin{algorithmic}[1]
	    \State \textbf{Input}: domain $\mathcal{X}$, GP prior with mean $\mu_0=0$ and the kernel function $k$, maximal iteration $N$
	    \State \textbf{Output}: Approximate maximizer $\mathbf{x}^{max}$
	    \State Let $\hat{\mathcal{D}}=\emptyset$
		\For {$t=1,2,\ldots N$}
		    \State Choose $\hat{\mathbf{x}}^{t} = \argmax_{\mathbf{x}\in\mathcal{X}}\hat{\mu}(\mathbf{x}\mid\hat{\mathcal{D}}=\{(\hat{\mathbf{x}}^{i},\hat{y}^{i})\}_{i=1}^{t-1}) + \sqrt{\beta_t}\hat{\sigma}(\mathbf{x}\mid\hat{\mathcal{D}}=\{(\hat{\mathbf{x}}^{i},\hat{y}^{i})\}_{i=1}^{t-1})$
		    \State Sample $\hat{y}^{t}=f\left(\hat{\mathbf{x}}^{t}\right) + \hat{\epsilon}_t $, where $\hat{\epsilon}_t\sim\mathcal{N}(0,\sigma_{0}^2)$ is a noise
		    \State Update $\hat{\mathcal{D}}=\hat{\mathcal{D}}\cup \{(\hat{\mathbf{x}}^{t},\hat{y}^{t})\}$
		    %\State Perform Bayesian update to obtain $\mu_t$ and $\sigma_t$.
			%\For {$actor=1,2,\ldots,N$}
				%\State Run policy $\pi_{\theta_{old}}$ in environment for $T$ time steps
				%\State Compute advantage estimates $\hat{A}_{1},\ldots,\hat{A}_{T}$
			%\EndFor
			%\State Optimize surrogate $L$ wrt. $\theta$, with $K$ epochs and minibatch size $M\leq NT$
			%\State $\theta_{old}\leftarrow\theta$
		\EndFor
		\State\Return $\hat{\mathbf{x}}^{max}$ that is equal to $\hat{\mathbf{x}}^{i}$ with maximal $\hat{y}^{i}$
	\end{algorithmic} 
	\label{alg:GP_UCB}
\end{algorithm}

Now we consider the standard GP-UCB algorithm (Algorithm~\ref{alg:GP_UCB}). Given the same GP prior and the same kernel function, since VS-GP-UCB and GP-UCB are two different algorithms, for each iteration they obtain different queries and have different posterior mean and standard deviation. We use the hat symbol to represent elements from GP-UCB, and in the following proof we use $\mu_{t}(\cdot)$ ($\hat{\mu}_{t}(\cdot)$) to represent $\mu(\cdot\mid\mathcal{D}=\{(\mathbf{x}^{i},y^{i})\}_{i=1}^{t})$ ($\hat{\mu}(\cdot\mid\hat{\mathcal{D}}=\{(\hat{\mathbf{x}}^{i},\hat{y}^{i})\}_{i=1}^{t})$) and $\sigma_{t}(\cdot)$ ($\hat{\sigma}_{t}(\cdot)$) to represent $\sigma(\cdot\mid\mathcal{D}=\{(\mathbf{x}^{i},y^{i})\}_{i=1}^{t})$ ($\hat{\sigma}(\cdot\mid\hat{\mathcal{D}}=\{(\hat{\mathbf{x}}^{i},\hat{y}^{i})\}_{i=1}^{t})$) for convenience. 

\citet{srinivas2009gaussian} has the following lemma:
\begin{lemma} [Lemma 5.6 in \citet{srinivas2009gaussian}]
    \label{lemma:5_6}
    Consider subsets $\mathcal{X}_t\subset\mathcal{X}$ with finite size, $|\mathcal{X}_t|<\infty$, pick $\delta\in (0,1)$ and set $\beta_{t}=2\log\left(\frac{\abs{\mathcal{X}_t}\pi_{t}}{\delta}\right)$, where $\sum_{t\geq 1}\pi_{t}^{-1}=1, \pi_t >0$. Running GP-UCB, we have that
    \begin{align*}
        \abs{f(\mathbf{x})-\hat{\mu}_{t-1}(\mathbf{x})}\leq \sqrt{\beta_t}\hat{\sigma}_{t-1}(\mathbf{x})\quad \forall \mathbf{x}\in\mathcal{X}_{t}, \forall t\geq 1
    \end{align*}
    holds with probability $\geq 1-\delta$.
\end{lemma}

By using the same proof procedure, we can easily obtain the following lemma: 

\begin{lemma} \label{lemma:5_6_VS}
    Consider subsets $\mathcal{X}_t\subset\mathcal{X}$ with finite size, $|\mathcal{X}_t|<\infty$, pick $\delta\in (0,1)$ and set $\beta_{t}=2\log\left(\frac{\abs{\mathcal{X}_t}\pi_{t}}{\delta}\right)$, where $\sum_{t\geq 1}\pi_{t}^{-1}=1, \pi_t >0$. Running VS-GP-UCB, we have 
    \begin{align*}
        \abs{f(\mathbf{x})-\mu_{t-1}(\mathbf{x})}\leq \sqrt{\beta_t}\sigma_{t-1}(\mathbf{x})\quad \forall \mathbf{x}\in\mathcal{X}_{t}, \forall t\geq 1
    \end{align*}
    holds with probability $\geq 1-\delta$.
\end{lemma}

Now consider $\mathcal{X}_{t}$ to be the following discretization: for each dimension $j=1,\dots , d$, discretize it with $\tau_t$ uniformly spaced points, and for each dimension $j=d+1,\dots , D$, discretize it with $\alpha\tau_t$ uniformly spaced points plus the j-th element in $\mathbf{x}^t$, $\mathbf{x}^t_j=\mathbf{x}^{0}_{j}$ (note that $\mathbf{x}^t_j$ is fixed when $j\geq d+1$). Denote $\widetilde{(\mathbf{x})}_t$ be the closest point in $\mathcal{X}_t$ to $\mathbf{x}$ under $L_1$ norm (note that $\mathbf{x}^t$ is the $t$-th query of the VS-GP-UCB algorithm, while $\widetilde{(\mathbf{x}^t)}_{t'}$ is its closet point in $\mathcal{X}_{t'}$), then because of the smoothness assumption, we have $w.p.\geq 1-\frac{\delta}{2}$:
\begin{align*}
     \abs{f(\mathbf{x})-f\left(\widetilde{(\mathbf{x})}_t\right)}\leq b\sqrt{\log\left(\frac{2Da}{\delta}\right)}\frac{d}{\tau_t} + \alpha b\sqrt{\log\left(\frac{2Da}{\delta}\right)}\frac{D-d}{\alpha\tau_t} &= b\sqrt{\log\left(\frac{2Da}{\delta}\right)}\frac{D}{\tau_t}, \\
     &\forall \mathbf{x}\in\mathcal{X}, \forall t\geq 1
\end{align*}

By choosing $\tau_t = Dt^{2}b\sqrt{\log\left(\frac{2Da}{\delta}\right)}$, we have $w.p.\geq 1-\frac{\delta}{2}$:
\begin{align*}
     \abs{f(\mathbf{x})-f(\widetilde{(\mathbf{x})}_t)}\leq\frac{1}{t^2}\quad\forall \mathbf{x}\in\mathcal{X}, \forall t\geq 1
\end{align*}

Under this situation, we have:
\begin{align*}
    \abs{\mathcal{X}_{t}}=(\alpha\tau_{t}+1)^{D-d}\tau_{t}^{d}=\left(\alpha Dt^{2}b\sqrt{\log\left(\frac{2Da}{\delta}\right)}+1\right)^{D-d}\left(Dt^{2}b\sqrt{\log\left(\frac{2Da}{\delta}\right)}\right)^{d}
\end{align*}
Combine with Lemma \ref{lemma:5_6_VS}, by setting:

\begin{align*}
    \beta_t=2\log\left(\frac{4\abs{\mathcal{X}_t}\pi_{t}}{\delta}\right)=2\log\frac{4\pi_t}{\delta} &+ 2(D-d)\log\left(\alpha Dt^{2}b\sqrt{\log\left(\frac{2Da}{\delta}\right)}+1\right) \\
    &+ 2d\log\left( Dt^{2}b\sqrt{\log\left(\frac{2Da}{\delta}\right)}\right)
\end{align*}
then w.p. $\geq 1-\delta$:
\begin{align*}
    \abs{f(\mathbf{x})-\mu_{t-1}\left(\widetilde{(\mathbf{x})}_t\right)} & \leq \abs{f(\mathbf{x}) - f\left(\widetilde{(\mathbf{x})}_t\right) + f\left(\widetilde{(\mathbf{x})}_t\right)-\mu_{t-1}\left(\widetilde{(\mathbf{x})}_t\right)} \\
    & \leq \abs{f(\mathbf{x}) - f\left(\widetilde{(\mathbf{x})}_t\right)} + \abs{f\left(\widetilde{(\mathbf{x})}_t\right)-\mu_{t-1}\left(\widetilde{(\mathbf{x})}_t\right)}\\
    & \leq \frac{1}{t^2} + \sqrt{\beta_t}\sigma_{t-1}\left(\widetilde{(\mathbf{x})}_t\right)\quad \forall \mathbf{x}\in\mathcal{X}, \forall t\geq 1.
\end{align*}

Likewise, by setting the same $\beta_{t}$, we have w.p. $\geq 1-\delta$:
\begin{align*}
    \abs{f(\mathbf{x})-\hat{\mu}_{t-1}\left(\widetilde{(\mathbf{x})}_t\right)} \leq \frac{1}{t^2} + \sqrt{\beta_t}\hat{\sigma}_{t-1}\left(\widetilde{(\mathbf{x})}_t\right)\quad \forall \mathbf{x}\in\mathcal{X}, \forall t\geq 1.
\end{align*}

Let $\mathbf{x}^{t}_{mix}=\{\hat{\mathbf{x}}^{t}_{[1:d]},\mathbf{x}_{[d+1:D]}^{0}\}$ (first $d$ variables are equal to those in $\hat{\mathbf{x}}^{t}$, and the others are equal to $\mathbf{x}_{[d+1:D]}^{0}$), again because of the smoothness assumption, we have $w.p. \geq 1-\frac{\delta}{2}$:

\begin{align*}
    \abs{f(\mathbf{x}^{t}_{mix})-f(\hat{\mathbf{x}}^{t})}\leq\alpha b\sqrt{\log\left(\frac{2Da}{\delta}\right)}\norm{\mathbf{x}^{0}_{[d+1:D]}-\hat{\mathbf{x}}^{t}_{[d+1:D]}}_{1}\leq \alpha b\sqrt{\log\left(\frac{2Da}{\delta}\right)}(D-d)
\end{align*}

Note that the point in $\mathcal{X}_{t}$ that is closest to $\mathbf{x}^{t}_{mix}$ is $\widetilde{(\mathbf{x}^{t}_{mix})}_{t}=\left\{\widetilde{(\hat{\mathbf{x}}^{t}_{[1:d]})}_{t},\mathbf{x}_{[d+1:D]}^{0}\right\}$ because all elements in $\mathbf{x}_{[d+1:D]}^{0}$ are contained in the discretization. Since $\mathbf{x}_{[1:d]}^{t}$ is the maximizer of UCB with fixed $D-d$ variables, we have:

\begin{align*}
    \mu_{t-1}\left(\widetilde{(\mathbf{x}^{t}_{mix})}_{t}\right) + \sqrt{\beta_t}\sigma_{t-1}\left(\widetilde{(\mathbf{x}^{t}_{mix})}_{t}\right) \leq \mu_{t-1}(\mathbf{x}^{t}) + \sqrt{\beta_t}\sigma_{t-1}(\mathbf{x}^{t})
\end{align*}

Similarly, considering GP-UCB algorithm, we have: 

\begin{align*}
    \hat{\mu}_{t-1}(\widetilde{(\mathbf{x}^{*})}_t) + \sqrt{\beta_t}\hat{\sigma}_{t-1}(\widetilde{(\mathbf{x}^{*})}_{t}) \leq \hat{\mu}_{t-1}(\hat{\mathbf{x}}^{t}) + \sqrt{\beta_t}\hat{\sigma}_{t-1}(\hat{\mathbf{x}}^{t})
\end{align*}

To finish the proof, we need to use another lemma in \citet{srinivas2009gaussian}:

\begin{lemma}[Lemma 5.5 in \citet{srinivas2009gaussian}]
\label{lemma:5_5}
    Let $\{\hat{\mathbf{x}}^{t}\}$ be a sequence of points chosen by GP-UCB, pick $\delta\in (0,1)$ and set $\beta_{t}=2\log\left(\frac{\pi_{t}}{\delta}\right)$, where $\sum_{t\geq 1}\pi_{t}^{-1}=1, \pi_{t}>0$. Then, 
    \begin{align*}
        \abs{f(\hat{\mathbf{x}}^{t})-\hat{\mu}_{t-1}(\hat{\mathbf{x}}^{t})}\leq \sqrt{\beta_{t}}\hat{\sigma}_{t-1}(\hat{\mathbf{x}}^{t}), \forall t\geq 1
    \end{align*}
    holds with probability $\geq 1-\delta$. 
\end{lemma}

By using the same proof procedure, we can easily obtain the following lemma: 

\begin{lemma}\label{lemma:5_5_VS}
    Let $\{\mathbf{x}^{t}\}$ be a sequence of points chosen by VS-GP-UCB, pick $\delta\in (0,1)$ and set $\beta_{t}=2\log\left(\frac{\pi_{t}}{\delta}\right)$, where $\sum_{t\geq 1}\pi_{t}^{-1}=1, \pi_{t}>0$. Then, 
    \begin{align*}
        \abs{f(\mathbf{x}^{t})-\mu_{t-1}(\mathbf{x}^{t})}\leq \sqrt{\beta_{t}}\sigma_{t-1}(\mathbf{x}^{t}), \forall t\geq 1
    \end{align*}
    holds with probability $\geq 1-\delta$. 
\end{lemma}

Finally, by using $\frac{\delta}{4}$ and setting (often set $\pi_t = \frac{\pi^{2}t^{2}}{6}$):
\begin{align*}
   \beta_{t}= 2\log\frac{16\pi_t}{\delta} + 2(D-d)\log\left(\alpha Dt^{2}b\sqrt{\log\left(\frac{8Da}{\delta}\right)}+1\right) + 2d\log\left( Dt^{2}b\sqrt{\log\left(\frac{8Da}{\delta}\right)}\right)
\end{align*}

$w.p.\geq 1-\delta$ the simple regret  $r_t$ of VS-GP-UCB, $r_t = f(\mathbf{x}^{*})-f(\mathbf{x}^{t}=\{\mathbf{x}_{[1:d]}^{t},\mathbf{x}_{[d+1:D]}^{0}\})$, is upper bounded by:
\begin{align*}
    r_t  & = f(\mathbf{x}^{*})-f(\mathbf{x}^{t}) \\
        %& = f(x^{*})-f(\hat{x}_{t})+f(\hat{x}_{t})-f([\hat{x}_{t}^{d},x_{t}^{D-d}])+f([\hat{x}_{t}^{d},x_{t}^{D-d}])-f([x_{t}^{d},x_{t}^{D-d}]) \\
        &\leq \hat{\mu}_{t-1}(\widetilde{(\mathbf{x}^{*})}_t) + \sqrt{\beta_t}\hat{\sigma}_{t-1}(\widetilde{(\mathbf{x}^{*})}_{t}) + \frac{1}{t^2} - f(\mathbf{x}^{t}) \\
        %& + \hat{\mu}_{t-1}(\widetilde{(x^{*})}_t) + \sqrt{\beta_t}\hat{\sigma}_{t-1}(\widetilde{(x^{*})}_{t}) + \frac{1}{t^2} - f([x_{t}^{d},x_{t}^{D-d}]) \\
        &\leq \hat{\mu}_{t-1}(\hat{\mathbf{x}}^{t}) + \sqrt{\beta_t}\hat{\sigma}_{t-1}(\hat{\mathbf{x}}^{t})+ \frac{1}{t^2} - f(\mathbf{x}^{t}) \\
        &=\hat{\mu}_{t-1}(\hat{\mathbf{x}}^{t}) + \sqrt{\beta_t}\hat{\sigma}_{t-1}(\hat{\mathbf{x}}^{t}) - f(\hat{\mathbf{x}}^{t}) + f(\hat{\mathbf{x}}^{t}) - f(\mathbf{x}^{t}_{mix})+ f(\mathbf{x}^{t}_{mix}) - f(\mathbf{x}^{t}) + \frac{1}{t^2} \\
        &\leq 2\sqrt{\beta_t}\hat{\sigma}_{t-1}(\hat{\mathbf{x}}^{t}) +\alpha b\sqrt{\log\left(\frac{8Da}{\delta}\right)}(D-d) +  \mu_{t-1}\left(\widetilde{(\mathbf{x}^{t}_{mix})}_{t}\right) \\
        &+ \sqrt{\beta_t}\sigma_{t-1}\left(\widetilde{(\mathbf{x}^{t}_{mix})}_{t}\right) - f(\mathbf{x}^{t}) + \frac{2}{t^2}\\ 
        &\leq 2\sqrt{\beta_t}\hat{\sigma}_{t-1}(\hat{\mathbf{x}}^{t}) +\alpha b\sqrt{\log\left(\frac{8Da}{\delta}\right)}(D-d) +  \mu_{t-1}(\mathbf{x}^{t}) + \sqrt{\beta_t}\sigma_{t-1}(\mathbf{x}^{t}) - f(\mathbf{x}^{t}) + \frac{2}{t^2}\\ 
        %&\leq \hat{\mu}_{t-1}(\hat{x}_{t}) + \sqrt{\beta_t}\hat{\sigma}_{t-1}(\hat{x}_{t}) -f(\hat{x}_{t}) + \mu_{t-1}([x^d_{t},x^{D-d}_{t}]) + \sqrt{\beta_t}\sigma_{t-1}([x^d_{t},x^{D-d}_{t}]) - f([x_{t}^{d},x_{t}^{D-d}]) \\
        %& + \frac{2}{t^{2}} + \alpha L(D-d) \\
        &\leq 2\sqrt{\beta_t}\hat{\sigma}_{t-1}(\hat{\mathbf{x}}^{t}) + 2\sqrt{\beta_t}\sigma_{t-1}(\mathbf{x}^{t}) + \frac{2}{t^{2}} + \alpha b\sqrt{\log\left(\frac{8Da}{\delta}\right)}(D-d)
\end{align*}
By using Lemma 5.4 in \citet{srinivas2009gaussian} and the Cauchy-Schwarz inequality, both $\sum_{t=1}^{N}2\sqrt{\beta_t}\hat{\sigma}_{t-1}(\hat{\mathbf{x}}^{t})$ and $\sum_{t=1}^{N}2\sqrt{\beta_t}\sigma_{t-1}(\mathbf{x}^{t})$ are upper bounded by $\sqrt{C_1 N\beta_{N}\gamma_{N}}$, where $C_1=\frac{8}{\log\left(1+\sigma_{0}^{-2}\right)}$, and $\gamma_{N}:=\max_{A\subset \mathcal{X}: |A|=N}\mathbf{I}(\mathbf{y}_{A};\mathbf{f}_{A})$ is the maximum information gain with a finite set of sampling points $A$, $\mathbf{f}_{A}=[f(\mathbf{x})]_{\mathbf{x}\in A}$, $\mathbf{y}_{A}=\mathbf{f}_{A} + \epsilon_{A}$. Therefore, we have:

\begin{align*}
    R_{N} = \sum_{t=1}^{N}r_{t}\leq 2\sqrt{C_1 N\beta_{N}\gamma_{N}} + \frac{\pi^{2}}{3} + \alpha b\sqrt{\log\left(\frac{8Da}{\delta}\right)}N(D-d)
\end{align*}

Hence:
\begin{align*}
    \frac{R_{N}}{N} = \frac{\sum_{t=1}^{N}r_{t}}{N}\leq 2\sqrt{C_1 \frac{\beta_{N}\gamma_{N}}{N}} + \frac{\pi^{2}}{3N} + \alpha b\sqrt{\log\left(\frac{8Da}{\delta}\right)}(D-d)
\end{align*}
%where $C_1=\frac{8}{\log\left(1+\sigma^{-2}\right)}$, and $\gamma_{T}:=\max_{A\subset D: |A|=T}\mathbf{I}(\mathbf{y}_{A};\mathbf{f}_{A})$ is the maximum information gain, $\mathbf{f}_{A}=[f(x)]_{x\in A}$, $\mathbf{y}_{A}=\mathbf{f}_{A} + \epsilon_{A}$. 

\end{proof}

\section{Detailed experimental settings and extended discussion of experimental results}

We use the framework of BoTorch to implement VS-BO. We compare VS-BO to the following existing BO methods: vanilla BO, which is implemented by the standard BoTorch framework\footnote{https://botorch.org}; REMBO and its variant REMBO Interleave~\citep{wang2016bayesian}, of which the implementations are based on \citet{metzen2016minimum}\footnote{https://github.com/jmetzen/bayesian\_optimization}; Dragonfly\footnote{https://github.com/dragonfly/dragonfly/}~\citep{kandasamy2020tuning}, which contains the Add-GP method; HeSBO~\citep{nayebi2019framework} which has already been implemented in Adaptive Experimentation Platform (Ax)\footnote{https://github.com/facebook/Ax/tree/master/ax/modelbridge/strategies}; And ALEBO\footnote{https://github.com/facebookresearch/alebo}~\citep{letham2020re}. By the time when we write this manuscript, source codes of the work \citet{spagnol2019bayesian} and \citet{eriksson2021high} have not been released, so we cannot compare VS-BO with these two methods. Both VS-BO and vanilla BO use Matern 5/2 as the kernel function and expected improvement as the acquisition function, and use limited-memory BFGS (L-BFGS) to fit GP and optimize the acquisition function. The number of initialized samples $N_{init}$ is set to $5$ for all methods, and $N_{vs}$ in VS-BO is set to $20$, $N_{is}$ is set to $10000$ for all experiments. The number of the interleaved cycle for REMBO Interleave is set to $4$. Since our algorithm aims to maximize the black-box function, all the test functions that have minimum points will be converted to the corresponding negative forms.

In synthetic experiments, as described in section 6.1, for each test function we add some unimportant variables as well as unrelated variables to make it high-dimensional. The standard Branin function $f_{Branin}$ has two dimensions with the input domain $\mathcal{X}_{Branin}=[-5,10]\times[0,10]$, and we construct a new Branin function $F_{branin}$ as the following:
\begin{align*}
    F_{branin}(\mathbf{x}) = f_{Branin}(\mathbf{x}_{[1:2]}) + 0.1f_{Branin}(\mathbf{x}_{[3:4]}) &+ 0.01f_{Branin}(\mathbf{x}_{[5:6]}), \\ 
    & \mathbf{x}\in \left(\bigotimes_{i=1}^{3}\mathcal{X}_{Branin}\right)\bigotimes_{i=1}^{44}[0,1]
\end{align*}
where $\bigotimes$ represents the direct product. We use $d_{e}=[2,2,2]$ to represent the dimension of the effective subspace of $F_{branin}$, the total effective dimension is $6$, however, the number of important variables is only $2$. 

Likewise, for the standard Hartmann6 function $f_{Hartmann6}$ that has six dimensions with the input domain $[0,1]^{6}$, we construct $F_{hm6}$ as:
\begin{align*}
    F_{hm6}(\mathbf{x}) = f_{Hartmann6}(\mathbf{x}_{[1:6]}) + 0.1f_{Hartmann6}(\mathbf{x}_{[7:12]}) + 0.01f_{Hartmann6}(\mathbf{x}_{[13:18]})\quad\mathbf{x}\in [0,1]^{50}
\end{align*}
and use $d_{e}=[6,6,6]$ to represent the dimension of the effective subspace. For the Styblinski-Tang4 function $f_{ST4}$ that has four dimensions with the input domain $[-5,5]^{4}$, we construct $F_{ST4}$ as: 
\begin{align*}
    F_{ST4}(\mathbf{x}) = f_{ST4}(\mathbf{x}_{[1:4]}) + 0.1f_{ST4}(\mathbf{x}_{[5:8]}) + 0.01f_{ST4}(\mathbf{x}_{[9:12]}) \quad\mathbf{x}\in [-5,5]^{50}
\end{align*}
and use $d_{e}=[4,4,4]$ to represent the dimension of the effective subspace. All synthetic experiments are run on the same Linux cluster that has 40 3.0 GHz 10-Core Intel Xeon E5-2690 v2 CPUs.

\begin{figure}[!ht]
  \centering
  %\fbox{\rule[-.5cm]{0cm}{4cm} \rule[-.5cm]{4cm}{0cm}}
  \includegraphics[width=0.99\textwidth]{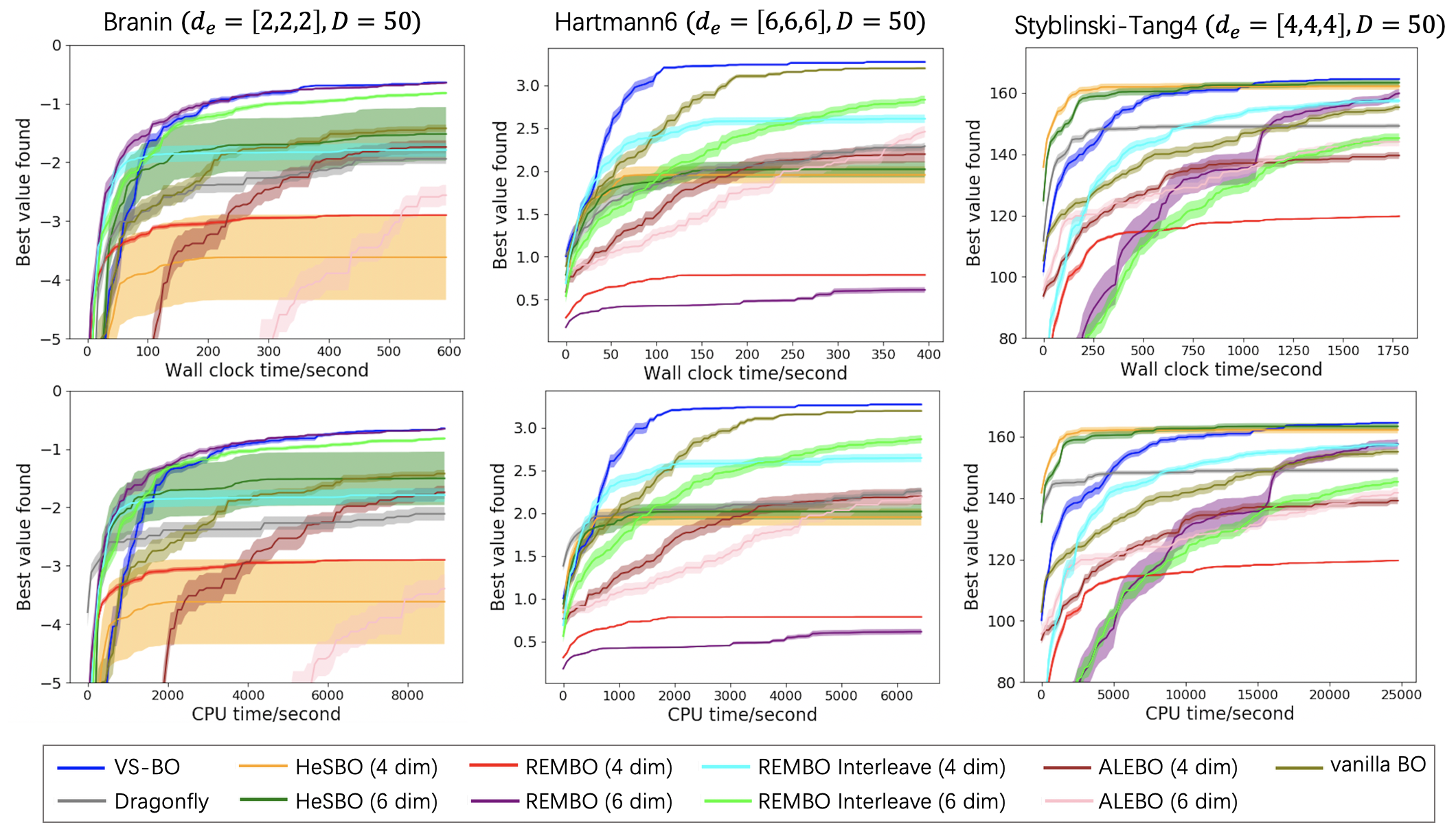}
  \caption{Performance of BO methods on Branin, Hartmann6 and Styblinski-Tang4 test functions. For each test function, we do 20 independent runs for each method. We plot the mean and 1/8 standard deviation of the best maximum value found by wall clock time used for BO (first row) and CPU time (second row).}
\label{fig:sythentic_result_time}
\end{figure}

\begin{figure}
  \centering
  %\fbox{\rule[-.5cm]{0cm}{4cm} \rule[-.5cm]{4cm}{0cm}}
  \includegraphics[width=0.99\textwidth]{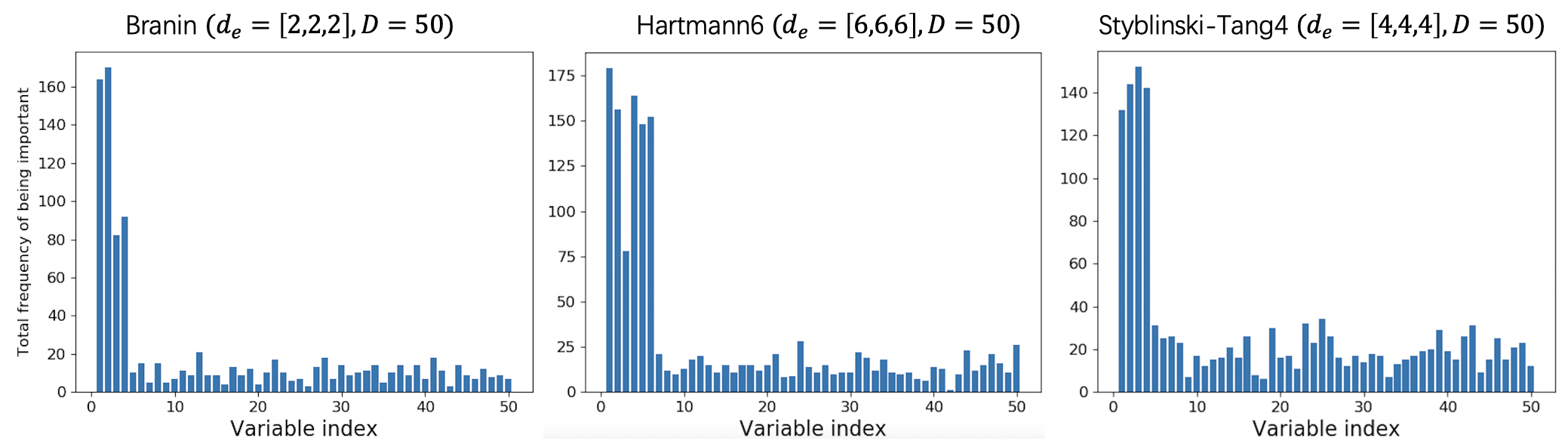}
  \caption{The total frequency of being chosen as important for each variable on Branin case (left), Hartmann6 case (middle) and Styblinski-Tang4 case (right). For the Branin function, the first two variables are important; for the Hartmann6 function, the first six variables are important; and for the Styblinski-Tang4 function, the first four variables are important.}
\label{fig:sythentic_f_chosen}
\end{figure}

Figure~\ref{fig:sythentic_result_iter} and ~\ref{fig:sythentic_result_time} show performances of different BO methods. They are compared under the fixed iteration budget (Figure~\ref{fig:sythentic_result_iter}), the fixed wall clock time budget used for BO itself (time for evaluating the black box function is excluded) or the fixed CPU time budget (Figure~\ref{fig:sythentic_result_time}). Figure~\ref{fig:sythentic_f_chosen} shows the frequency of being chosen as important for each variable in steps of variable selection of VS-BO. Since we do 20 runs of VS-BO on each test function, each run has $210$ iterations and important variables are re-selected every $20$ iterations, the maximum frequency each variable can be chosen as important is $200$. In the Branin case with $d_e=[2,2,2]$, the first two variables are important, and the left panel of Figure~\ref{fig:sythentic_f_chosen} shows that the frequency of choosing the first two variables as important is significantly higher than that of choosing the other variables; In the Hartmann6 case with $d_e=[6,6,6]$, the first six variables are important, and the middle panel of Figure~\ref{fig:sythentic_f_chosen} shows that the frequency of choosing the first six variables as important is the highest; In the Styblinski-Tang4 case with $d_e=[4,4,4]$, the first four variables are important, and again the right panel of Figure~\ref{fig:sythentic_f_chosen} shows that the frequency of choosing the first four variables as important is the highest. These results indicate that VS-BO is able to find real important variables and control false positives. 

\begin{figure}
  \centering
  %\fbox{\rule[-.5cm]{0cm}{4cm} \rule[-.5cm]{4cm}{0cm}}
  \includegraphics[width=0.99\textwidth]{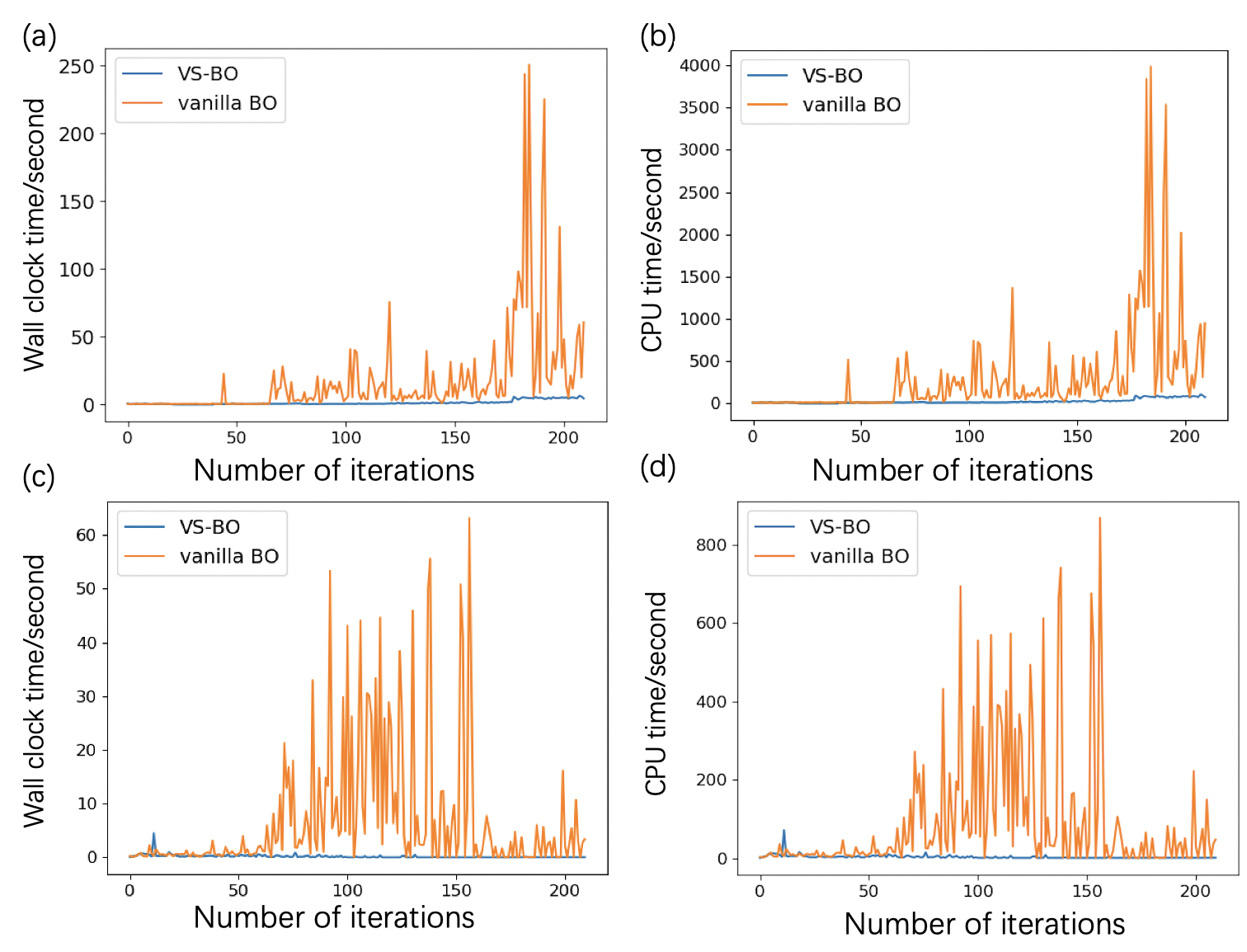}
  \caption{The wall clock time or CPU time comparison between VS-BO and vanilla BO for each iteration. The Branin test function with $d_{e}=[2,2,2]$ and $D=50$ is run here. (a) Wall clock time comparison at the GP fitting step (b) CPU time comparison at the GP fitting step (c) Wall clock time comparison at the acquisition function optimization step (d) CPU time comparison at the acquisition function optimization step}
\label{fig:runtime_compare_branin}
\end{figure}

Figure~\ref{fig:runtime_compare_branin} shows the wall clock time or CPU time comparison between VS-BO and vanilla BO for each iteration under the step of fitting a GP or optimizing the acquisition function. As the number of iterations increases, the runtime of vanilla BO increases significantly, while for VS-BO the runtime only has a slight increase. These results empirically show that VS-BO is able to reduce the runtime of BO process.

\begin{figure}[!ht]
  \centering
  %\fbox{\rule[-.5cm]{0cm}{4cm} \rule[-.5cm]{4cm}{0cm}}
  \includegraphics[width=0.89\textwidth]{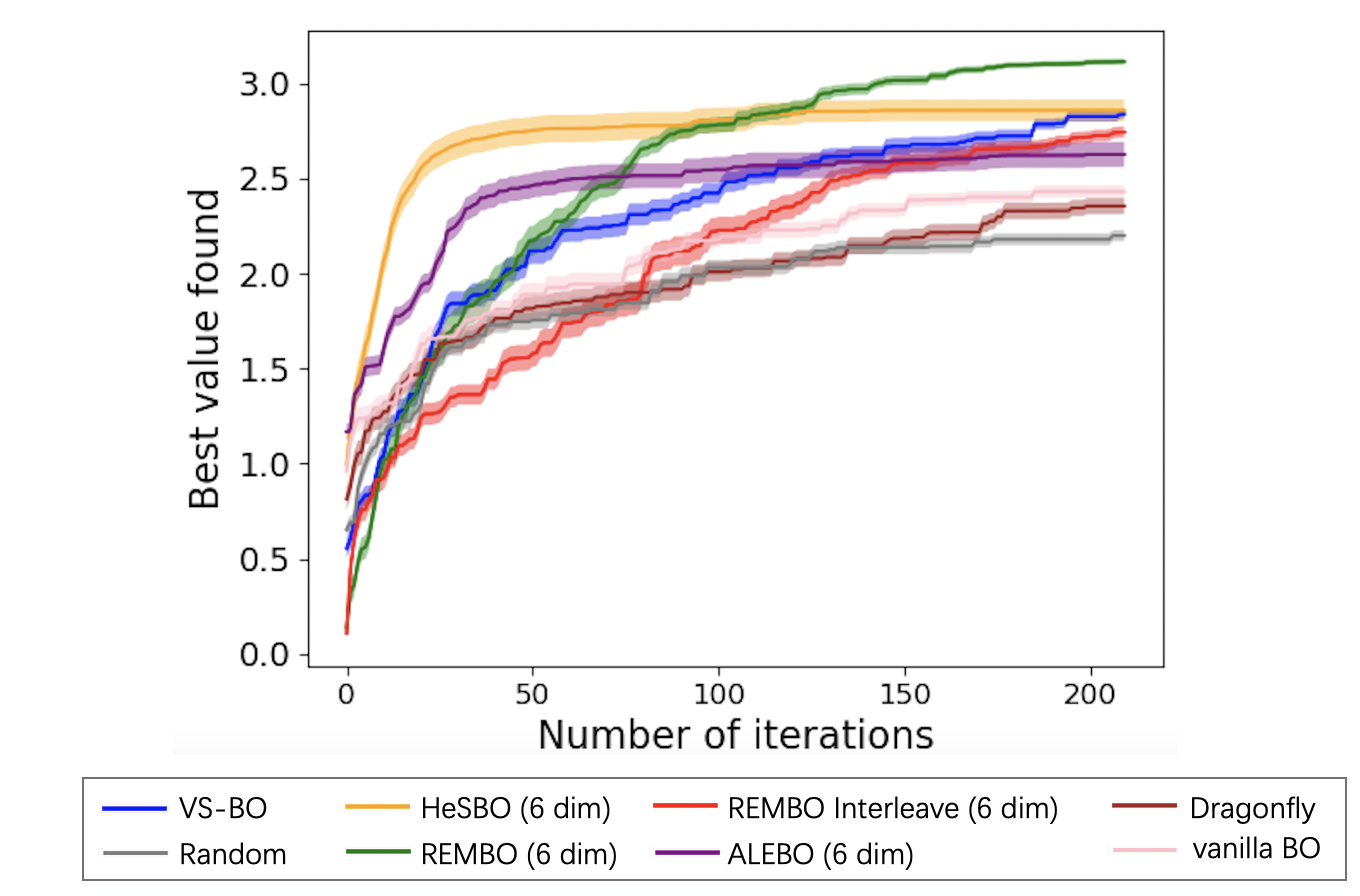}
  \caption{Performance of BO methods on the rotational Hartmann6 function. We do 20 independent runs for each method. We plot the mean and 1/8 standard deviation of the best maximum value found by iterations.}
\label{fig:rotation_compare_iter}
\end{figure}

To test the performance of VS-BO on the function that has a non-axis-aligned subspace, we construct a rotational Hartmann6 function $F_{rot\_H}(\mathbf{x})$ by using the same way as \citet{letham2020re}. We sample a rotation matrix $A$ from the Haar distribution on the orthogonal group $SO(100)$~\citep{stewart1980efficient}, and take the form of $F_{rot\_H}(\mathbf{x})$ as the following:
\begin{align*}
    F_{rot\_H}(\mathbf{x})=f_{Hartmann6}(A[:6]\mathbf{x})
\end{align*}
where $A[:6]\in\mathbb{R}^{6\times 100}$ represents the first $6$ rows of $A$ and $\mathbf{x}\in [-1,1]^{100}$ is the input. We then run VS-BO as well as other methods on this function and compare their performance. Figure~\ref{fig:rotation_compare_iter} shows that in this case REMBO with $d=6$ has the best performance. VS-BO also has a good performance, and surprisingly it outperforms several embedding-based methods such as ALEBO ($d=6$) and REMBO Interleave ($d=6$), although it tries to learn an axis-aligned subspace. 

%VS-BO outperforms ALEBO and random search in this case, indicating that our method is also able to maximize the function that has a non-axis-aligned subspace. 

\begin{figure}[!ht]
  \centering
  %\fbox{\rule[-.5cm]{0cm}{4cm} \rule[-.5cm]{4cm}{0cm}}
  \includegraphics[width=0.99\textwidth]{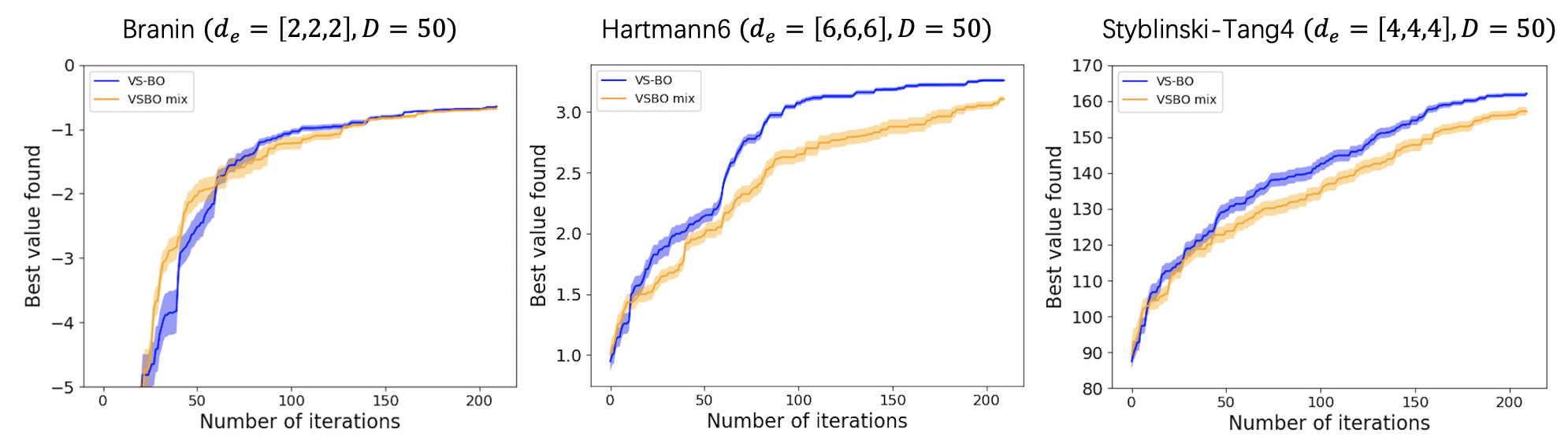}
  \caption{Performance of VS-BO and VSBO mix on Branin, Hartmann6 and Styblinski-Tang4 test functions. For each test function, we do 20 independent runs for each method. We plot the mean and 1/8 standard deviation of the best maximum value found by iterations.}
\label{fig:VSBO_variant_comp}
\end{figure}

\citet{spagnol2019bayesian} introduces several methods to sample unimportant variables and shows that the mix strategy is the best, that is for each iteration, the algorithm samples values of unimportant variables from the uniform distribution with probability $0.5$ or use the values from the previous query that has the maximal function value with probability $0.5$. To compare the mix strategy with our sampling strategy, i.e. sampling from CMA-ES related posterior, we replace our strategy with the mix strategy in VS-BO, creating a variant method called VSBO mix. All the other parts in VSBO mix are the same as those in VS-BO. We compare VS-BO with VSBO mix on these three synthetic functions, and Figure~\ref{fig:VSBO_variant_comp} show that in general our sampling strategy is clearly better than the mix strategy. 

For real-world problems, the rover trajectory problem is a high-dimensional optimization problem with input domain $[0,1]^{60}$. The problem setting in our experiment is the same as that in \citet{wang2017batched}, and the source code of this problem can be found in https://github.com/zi-w/Ensemble-Bayesian-Optimization. MOPTA08 is another high-dimensional optimization problem with input domain $[0,1]^{124}$. It has one objective function $f_{mopta}(\mathbf{x})$ that needs to be minimized and $68$ constraints $c_{i}(\mathbf{x}), i\in\{1,2,\ldots 68\}$. Similar to \citet{eriksson2021high}, we convert these constraints to soft penalties and convert the minimization problem to the maximization problem by adding a minus at the front of the objective function, i.e., we construct the following new function $F_{mopta}$:
\begin{align*}
    F_{mopta}(\mathbf{x}) = -\left(f_{mopta}(\mathbf{x}) + 10\sum_{i=1}^{68}\max (0,c_{i}(\mathbf{x}))\right)
\end{align*}
The Fortran codes of MOPTA can be found in https://www.miguelanjos.com/jones-benchmark and we further use codes in https://gist.github.com/denis-bz/c951e3e59fb4d70fd1a52c41c3675187 to wrap up it in python. All experiments for these two real-world problems are run on the same Linux cluster that has 80 2.40 GHz 20-Core Intel Xeon 6148 CPUs.

\begin{figure}
  \centering
  %\fbox{\rule[-.5cm]{0cm}{4cm} \rule[-.5cm]{4cm}{0cm}}
  \includegraphics[width=0.99\textwidth]{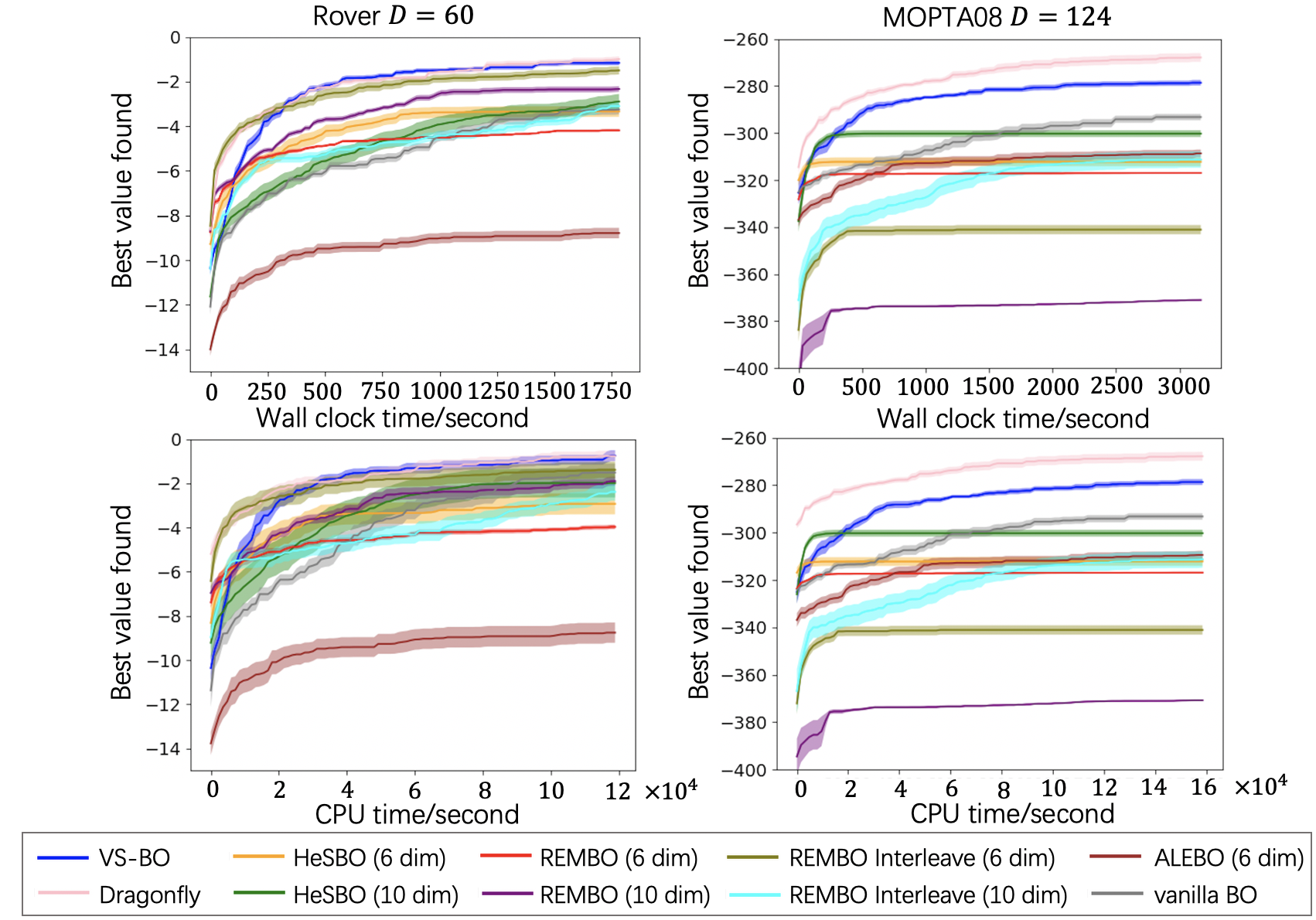}
  \caption{Performance of BO methods on the rover trajectory and MOPTA08 problems. We do 20 independent runs on the rover trajectory problem and 15 on the MOPTA08 problem. We plot the mean and 1/4 standard deviation of the best maximum value found by wall clock time (first row) and CPU time (second row).}
\label{fig:real_result_time}
\end{figure}

\begin{figure}
  \centering
  %\fbox{\rule[-.5cm]{0cm}{4cm} \rule[-.5cm]{4cm}{0cm}}
  \includegraphics[width=0.99\textwidth]{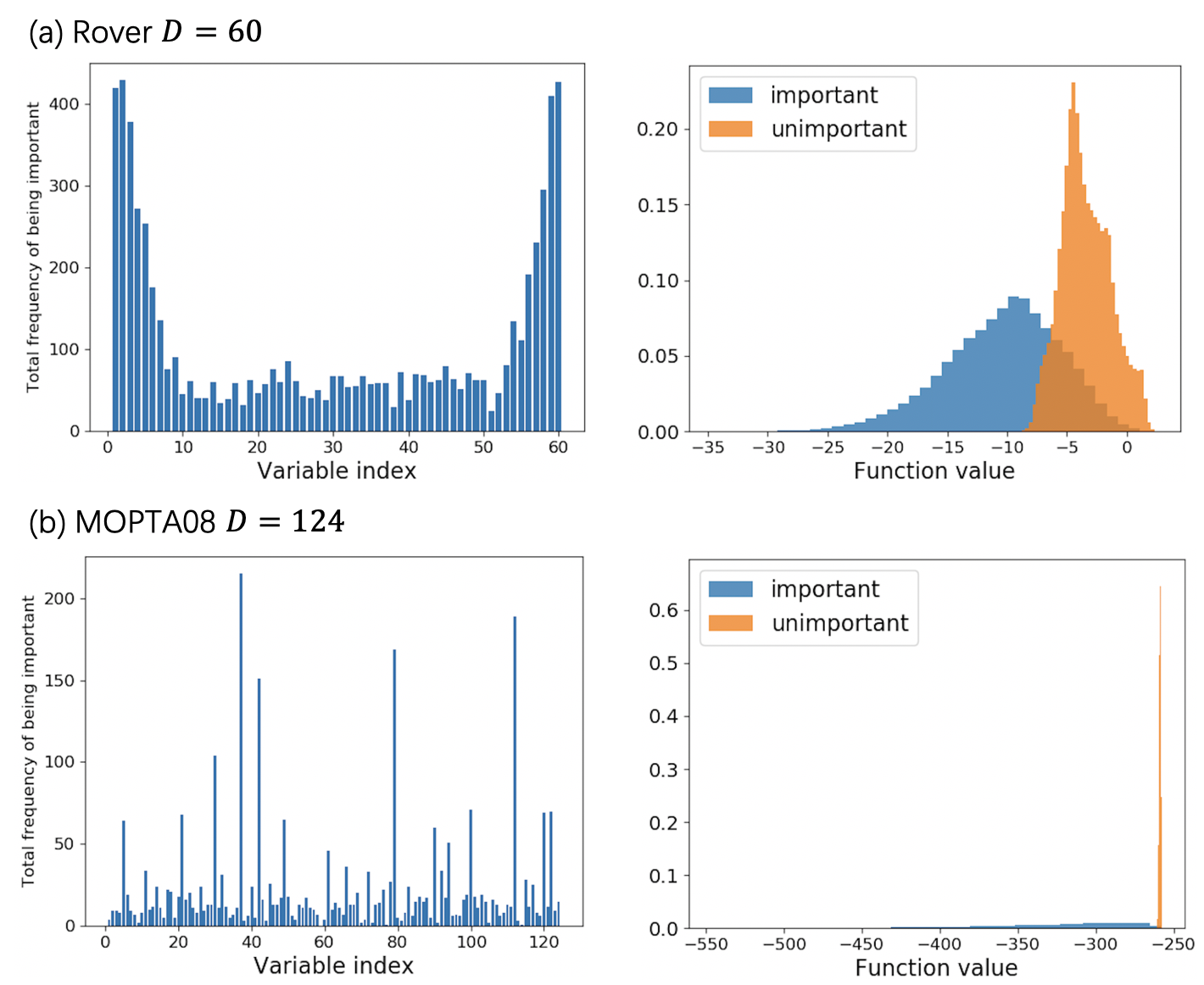}
  \caption{(left column) The total frequency of being chosen as important for each variable on the rover trajectory and MOPTA08 problems. (right column)The distribution of function values when sampling the first $5$ variables that have been chosen most frequently (important) or the first $5$ variables that have been chosen least frequently (unimportant) with all the other variables fixed.}
\label{fig:real_F_permutation}
\end{figure}

Figure~\ref{fig:real_result_iter} and ~\ref{fig:real_result_time} show performances of different BO methods on these two real-world problems. Note that Dragonfly performs quite well if the wall clock or CPU time used for BO is fixed, but not as good as VS-BO under the fixed iteration budget. Similar to Figure~\ref{fig:sythentic_f_chosen}, the left column of Figure~\ref{fig:real_F_permutation} shows the frequency of being chosen as important for each variable in steps of the variable selection of VS-BO. As described in section 6.2, we design a sampling experiment to test the accuracy of the variable selection. The indices of the first $5$ variables that have been chosen most frequently are $\{1,2,3,59,60\}$ on the rover trajectory problem and $\{30,37,42,79,112\}$ on MOPTA08, and the indices of the first $5$ variables that have been chosen least frequently are $\{15,18,29,38,51\}$ and $\{59,77,91,105,114\}$ respectively. The total number of samples in each set is $800000$.  The right column of Figure~\ref{fig:real_F_permutation} shows the empirical distributions of function values from two sets of samples. The significant difference between two distributions in each panel tells us that changing the values of variables that have been chosen more frequently can alter the function value more significantly, indicating that these variables are more important. 

\citet{eriksson2021high} shows an impressive performance of their method, SAASBO, on MOPTA08 problem. According to their results, SAASBO outperforms VS-BO in this case. We don't compare VS-BO with SAASBO in our work since the source code of SAASBO has not been released. One potential drawback of SAASBO is that it is very time consuming. For each iteration, SAASBO needs significantly more runtime than ALEBO (section C of \citet{eriksson2021high}), while our experiments show that ALEBO is a method that is significantly more time-consuming than VS-BO, sometimes even more time-consuming than vanilla BO, especially when $d$ is large (for example when $d\geq 10$). Therefore, SAASBO might not be a good choice for the case when the runtime of BO needs to be considered. 

%Optionally include extra information (complete proofs, additional experiments and plots) in the appendix.
%This section will often be part of the supplemental material.

\end{document}